\pdfoutput=1

\documentclass[11pt]{article}

\usepackage[preprint]{acl}

\usepackage{times}
\usepackage{latexsym}

\usepackage[T1]{fontenc}

\usepackage[utf8]{inputenc}

\usepackage{microtype}

\usepackage{inconsolata}

\usepackage{graphicx}

\newcommand{\rparagraph}[1]{\vspace{1.2mm}\noindent\textbf{#1.}}

\newcommand{\sparagraph}[1]{\vspace{0.0mm}\noindent\textbf{#1.}}

\newcommand{\sparagraphnodot}[1]{\vspace{0.0mm}\noindent\textbf{#1}}

\newcommand{\xlt}{{\textsc{xlt}}\xspace}
\newcommand{\zsxlt}{{\textsc{zs-xlt}}\xspace}
\newcommand{\ttrain}{{\textsc{TTrain}}\xspace}
\newcommand{\ttest}{{\textsc{TTest}}\xspace}
\newcommand{\sdev}{{\textsc{s-dev}}\xspace}
\newcommand{\tdev}{{\textsc{t-dev}}\xspace}
\newcommand{\llmvec}{{\texttt{LLM2Vec}}\xspace}
\usepackage{hyperref}

\usepackage{xspace}
\usepackage{booktabs}
\usepackage{amsmath}
\usepackage{amssymb}
\usepackage{graphicx}
\usepackage{adjustbox}
\usepackage{multirow}
\usepackage{comment}
\usepackage{color, colortbl}
\usepackage{arydshln}
\usepackage{hhline}
\usepackage{array}
\usepackage{makecell}

\title{Self-Distillation for Model Stacking Unlocks \\ Cross-Lingual NLU in 200+ Languages}

\author{Fabian David Schmidt\textsuperscript{1}, Philipp Borchert\textsuperscript{2}, Ivan Vulić\textsuperscript{3}, Goran Glavaš\textsuperscript{1} \\
  \textsuperscript{1} Center For Artificial Intelligence and Data Science, University of Würzburg, Germany \\
  \textsuperscript{2} IESEG School of Management, France; KU Leuven, Belgium\\
  \textsuperscript{3} Language Technology Lab, University of Cambridge, United Kingdom \\
  \texttt{\{fabian.schmidt, goran.glavas\}@uni-wuerzburg.de} \\
  \texttt{philippniklas.borchert@kuleuven.be}\\
  \texttt{iv250@cam.ac.uk} }

\begin{document}
\maketitle
\begin{abstract}
LLMs have become a go-to solution not just for text generation, but also for natural language understanding (NLU) tasks. Acquiring extensive knowledge through language modeling on web-scale corpora, they excel on English NLU, yet struggle to extend their NLU capabilities to underrepresented languages. 
In contrast, machine translation models (MT) produce excellent multilingual representations, resulting in strong translation performance even for low-resource languages. MT encoders, however, lack the knowledge necessary for comprehensive NLU that LLMs obtain through language modeling training on immense corpora. 
In this work, we get the best both worlds by integrating MT encoders directly into LLM backbones via sample-efficient self-distillation. The resulting MT-LLMs preserve the inherent multilingual representational alignment from the MT encoder, allowing lower-resource languages to tap into the rich knowledge embedded in English-centric LLMs.
Merging the MT encoder and LLM in a single model, we mitigate the propagation of translation errors and inference overhead of MT decoding inherent to discrete translation-based cross-lingual transfer (e.g., translate-test). Evaluation spanning three prominent NLU tasks and 127 predominantly low-resource languages renders MT-LLMs highly effective in cross-lingual transfer. MT-LLMs substantially and consistently outperform translate-test based on the same MT model, showing that we truly unlock multilingual language understanding for LLMs. 

  



\end{abstract}

\section{Introduction}
Large Language Models (LLMs) have become the swiss-army knife for natural language understanding (NLU) in English.
When pretrained with language modelling on trillions of tokens, LLMs excel at complex NLU tasks with minimal or no labeled data \cite{gpt3,touvron2023llama,llama3modelcard}. 
Although these models are predominantly trained on English texts, typically comprising more than 80\% of their training data \cite{touvron2023llama, llama3modelcard, aryabumi2024aya}, they show strong NLU capabilities also in other high-resource languages~\cite{blevins2022language,zhu2023extrapolating}.\footnote{For instance, 5\% of the 15T pretraining dataset of Llama 3 comprise non-English data spanning over 30 languages. Aya is tailored for NLU across 23 high-resource languages.} 
However, LLM performance degrades in cross-lingual transfer to languages that are typologically distant from English or virtually \textit{unseen} at pretraining \cite{ojo2024good,holtermann2024evaluating,razumovskaia2024analyzing}. This performance degradation restricts the effectiveness of LLMs primarily to English and a tiny subset of high-resource languages and underscores shortcomings in their adaptability to underrepresented low-resource languages, thereby amplifying the cross-lingual language technology gap~\cite{joshi-etal-2020-state,razumovskaia2024analyzing}.

\begin{figure}
    \centering
    \includegraphics[scale=0.12]{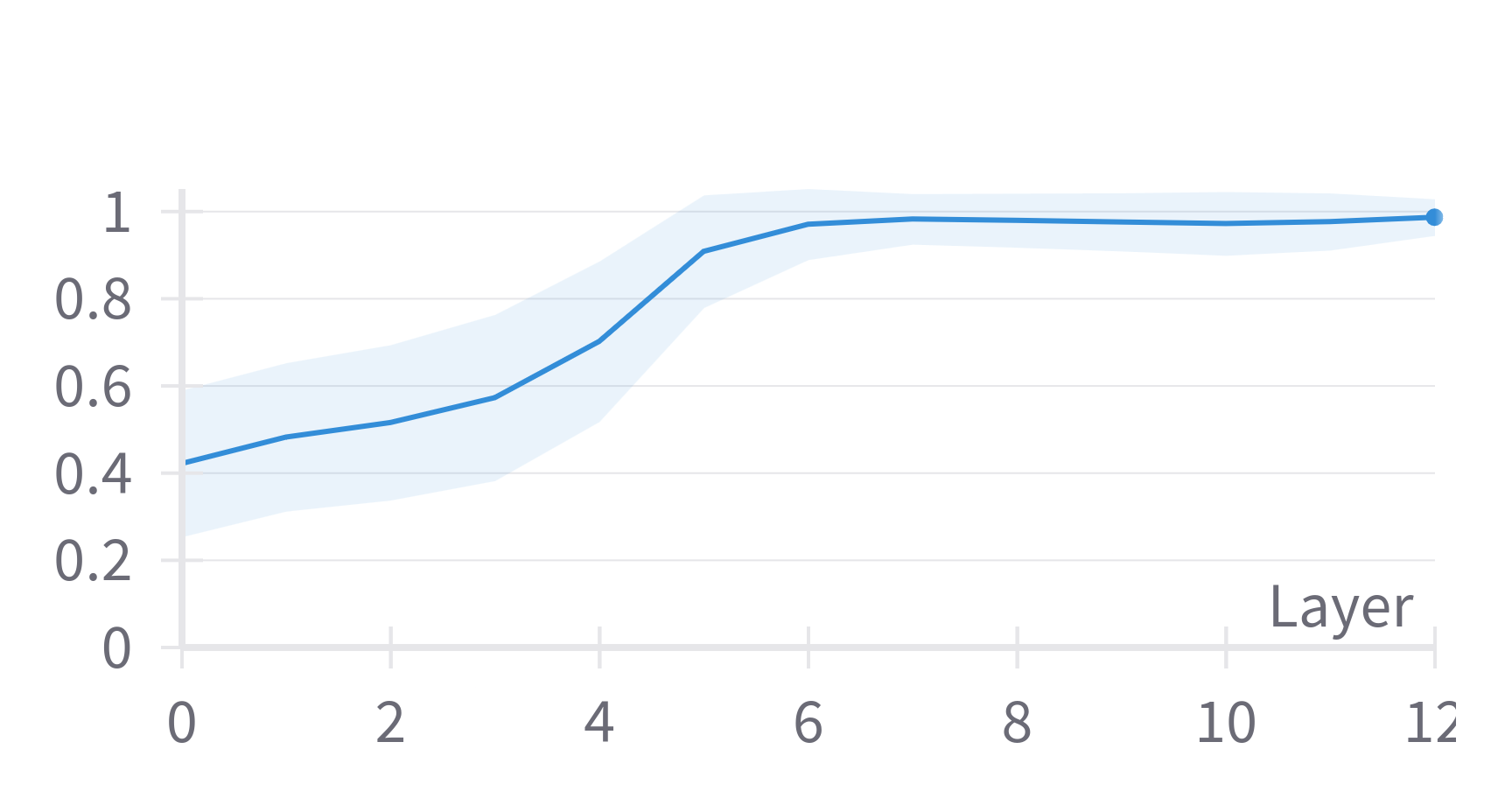}
    \hspace{3cm} 
    \vspace{-2.25em} 
    \caption{Mean \& std. dev. of sentence translation retrieval accuracy with \texttt{BERTScore} for NLLB encoder outputs on pooled \texttt{dev} \& \texttt{devtest} sets of \textsc{Flores200} by layer \cite{nllbteam2022language}.}
    \label{fig:nllb-flores-bert-score}
\vspace{-4mm}
\end{figure}

In contrast, publicly available \textit{machine translation} models like NLLB~\cite{nllbteam2022language} and MADLAD-400~\cite{kudugunta2023madlad400} are by design oriented towards and showcase ever more inclusiveness; they provide some machine translation capabilities between more than 200 and 400 languages, respectively, in any language direction. 
Unlike LLMs, machine translation (MT) models, and specifically MT encoders, are designed to semantically align textual representations in a unified embedding space, as demonstrated by their sentence retrieval performance on the \textsc{Flores200} dataset (cf. Figure \ref{fig:nllb-flores-bert-score}). 

However, the MT models lack various types of knowledge (e.g., world knowledge, commonsense knowledge), commonly acquired through large-scale language modeling pre-training. Consequently, in multilingual NLU tasks and respective cross-lingual transfer they underperform even smaller multilingual encoders like XLM-R \cite{conneau-etal-2020-unsupervised} (see Appendix \ref{subsec:nllb-vs-xlmr} for an empirical comparison). Because of this, MT models are typically used fully downstream, to translate training and/or test data from the source to the target languages, extending the wide availability of task-annotated English corpora to the target languages~\cite{ruder-etal-2021-xtreme,artetxe-etal-2023-revisiting,ebing2023translate}. 
Translating training data (\ttrain) involves substantial computational resources but yields strong \xlt performance. 
\ttrain nevertheless requires LLMs to support the target languages, which does not hold true for low-resource languages \cite{ojo2024good}. 
Translating test data, on the other hand, enables \zsxlt with monolingual LMs, but it incurs an additional inference overhead from MT and generally offers performance that is slightly inferior to \ttrain. Both \ttrain and \ttest aim to align the input to accommodate the shortcomings of the LLM representation space, resorting for this to discrete natural language translations coming from the MT decoder. 
These methods fail to preserve the rich latent representations from the MT encoder and propagate translation errors to LLMs, thereby reducing downstream performance \cite{ponti2021modelling}.

In this work, we thus propose to merge MT encoders directly with LLMs, creating a unified multilingual LLM for enhanced cross-lingual NLU, termed MT-LLM. The merger of the two models unlocks the potential to combine \textbf{1)} the general knowledge available in the original LLM for English and a handful of high-resource languages and \textbf{2)} powerful multilingual representations and their cross-lingual semantic alignment available in the MT encoder (see Figure \ref{fig:nllb-flores-bert-score}). The key idea involves enabling the LLM to directly integrate the output representations from MT encoders, this way extending its NLU performance to virtually all languages supported by the MT encoder. 

We align MT encoders with LLMs via self-distillation in two steps. The objective in the first, self-supervised adaptation step is sequence-level alignment between the original LLM and the MT-equipped LLM (MT-LLM). Second, we then address the distributional shifts inherent to adaptation from general-purpose data to downstream task data through task-specific self-distillation. We fine-tune the LLM on labeled task data, then transfer this task knowledge to the MT-LLM by aligning the task-specific output representations.

\rparagraph{Contributions}
\textbf{1)} To the best of our knowledge, we are the first to successfully integrate MT encoders into language model backbones for \xlt, thereby enabling \zsxlt to all languages supported by the MT encoder. This integration yields two key benefits: \zsxlt performance consistently improves over \ttest, while simultaneously reducing inference cost by eliminating the need to translate test instances. In turn, we show that the integration is highly efficient and only requires a few self-supervised adaptation steps to yield performance improvements over the LLM backbone.
\textbf{2)} We empirically show that our approach is agnostic to different types of LLM backbones, i.e., it improves the \zsxlt capabilities of both decoder-only and encoder-only models.
\textbf{3)} We compare \zsxlt and \ttest extensively and \textit{fairly} on a range of tasks and a wide spectrum of (all supported) languages.\footnote{Our unified MT-LLM approach integrates additional MT encoder parameters, while \ttest utilizes both the MT encoder and decoder for translating test instances into English. Additionally, \zsxlt is commonly evaluated on languages unsupported by the LLM, where MT models are employed to bridge this gap in both \ttest and \ttrain.} Unlike existing work, we make sure that both cross-lingual transfer approaches---latent with MT-LLM and discrete with \ttest---are evaluated on an equal footing. Our results demonstrate that \zsxlt with MT-LLM surpasses \ttest on NLU tasks when both rely on the same MT model.

\section{Related Work}
\sparagraphnodot{Translation-based \xlt}
is a strong \xlt baseline \cite{ruder-etal-2021-xtreme,ebrahimi-etal-2022-americasnli,aggarwal-etal-2022-indicxnli}.
Previous studies have explored various techniques for leveraging translated training data in \xlt (\ttrain): these include training on translated data in a single target language \cite{ebrahimi-etal-2022-americasnli}, using concatenated data from all target languages \cite{ruder-etal-2021-xtreme}, sequential training starting with the source language followed by the translated target language \cite{aggarwal-etal-2022-indicxnli}, and jointly training on both combined \cite{chen-etal-2023-frustratingly}. Recent studies have also benchmarked translating test data (\ttest) \cite{hu2020xtreme, isbister-etal-2021-stop}, which enables \zsxlt without the need for extensive fine-tuning for each target language, as in the case of \ttrain.
Moreover, both paradigms can be combined by training on round-trip translated noisy source data (translating source-language data to the target language and back) and evaluating on target language test data translated to the source \cite{oh-etal-2022-synergy,artetxe-etal-2023-revisiting,ebing2023translate}. 
Translating training or test data is essentially a \textit{discrete} approach for adjusting the input (i.e., its language) to the LLM (i.e., language that the LLM is proficient in). In contrast, we propose to align latent representation of input, produced by the MT encoders, to the representation space of the LLM backbone via self-distillation, effectively bypassing translation errors that arise from the discrete translation, output of the MT decoder. 
By retaining continuous MT encoder representations and avoiding their discretization in the MT decoder, our approach also reduces the time and cost of inference vis-a-vis \ttest. This also means that the MT-LLMs (unlike English-centric LLMs) can also reap further gains from \ttrain, particularly for low-resource languages unseen during pretraining.

While few studies investigated the integration of rich MT representations into LMs, these efforts have generally focused on task-specific integration, without achieving a global representation alignment between the MT encoder and the (large) language model \cite{ponti2021modelling, unanue-etal-2023-t3l}. Our approach addresses this limitation by achieving task-agnostic representation alignment between MT and LM before task specialization.

\rparagraph{Cross-lingual Transfer with LLMs} Widely used LLMs are predominantly trained on English data with English text accounting for 80-90\% of their pretraining corpora \cite{touvron2023llama,llama3modelcard}. Despite this imbalance, LLMs demonstrate a surprisingly strong performance in (high-resource) languages, which account for only a small fraction of their pretraining corpora \cite{blevins2022language}. The pretraining focus on English limits the NLU capabilities of LLMs in many low(er)-resource languages, and languages linguistically distant from English~\cite{ojo2024good}. Various methods adapt LLMs to languages not covered during pretraining, including continued pretraining \cite{shliazhko2023mgpt,fujii2024continual}, self-instruction \cite{wei2023polylm}, and vocabulary extension \cite{zhao2024llama}. These methods yield gains in model's target language generation capabilities; however, recent work shows that better generation does not translate to stronger NLU performance \cite{razumovskaia2024analyzing}.



\section{Methodology}
\begin{figure*}[t!]
    \centering
    \includegraphics[width=0.95\textwidth]{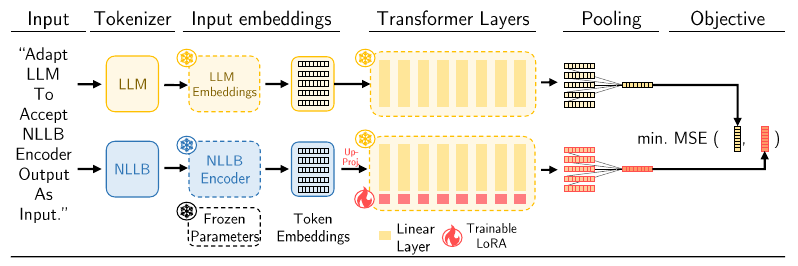}
    \vspace{-1mm}
    \caption{Overview of Stage 1: we merge an MT encoder (NLLB, as a representative MT model) with an LLM (obtaining MT-LLM). We train, in a self-supervised distillation setup, the up-projection and LoRA adapters of the MT-LLM by forcing its output to match (via mean-squared error) the output of the LLM itself.}
    \label{fig:nllb-llm2vec-architecture}
\vspace{-0.3cm}
\end{figure*}
\begin{figure}[!t]
    \centering
    \includegraphics[width=0.99\linewidth]{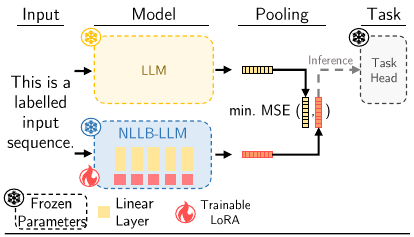}
    \caption{Overview of the architecture in Stage 2: task-specific distillation (again assuming the use of NLLB).}
    \label{fig:nllb-llm2vec-task-distillation}
\vspace{-0.5cm}
\end{figure}

\sparagraph{Idea in a Nutshell}
Moving beyond translation-based \xlt at the discrete (input data) level, we propose a method that merges a base LLM and an MT encoder into a massively multilingual \textit{`MT-LLM'}. This integration enables the MT-LLM model to perform \zsxlt to any language supported by the MT encoder by leveraging its multilingual language alignment capabilities that substantially extend those of the base LLM (see again Figure \ref{fig:nllb-flores-bert-score}).

More concretely, we hypothesize that by fine-tuning additional modular parameters on top of the base LLM, we can align the output representations of the merged MT-LLM with the original output representations of the base LLM. In other words, we learn the MT-LLM alignment via distillation with the LLM itself as the teacher. As a result, the MT encoder representations, which are of high-quality for many languages, act as input for the merged MT-LLM. The MT-LLM merge happens in two stages: \textbf{1)} self-supervised (general, task-agnostic) adaptation and \textbf{2)} task-specific distillation; we describe both in what follows.

\rparagraph{Stage 1: Self-Supervised General Adaptation} 
Figure~\ref{fig:nllb-llm2vec-architecture} illustrates the approach in the first stage. We `vertically' fuse a multilingual MT encoder $E$ and an LLM $M$ into a multilingual MT-LLM $E \times M$. To this end, we introduce two sets of new, trainable parameters $\Theta$: We first initialize a projection $U \in \mathbb{R}^{d_E \times d_M}$ that maps the output representation space $\mathbb{R}^{d_E}$ of the MT encoder $E$ into the input embedding space $\mathbb{R}^{d_M}$ of the LLM $M$. We then insert low-rank adapters (LoRAs)~\cite{hu2022lora} $\Delta W_{i=1}^{|W|}$ into the linear layers $W$ of the LLM $M$. All other parameters of the MT encoder $E$ and the LLM $M$ are frozen.

The principal idea is to train the new modular parameters $\Theta=\{U, \Delta W_{i=1}^{|W|}\}$ to enable the LLM backbone $M$ to `understand' output token embeddings of the massively multilingual MT encoder.
To this end, we utilize the original LLM as a teacher, which guides the self-supervised training process of our stacked MT-LLM. In the initial pass, we feed the input sequence $S$ through the original LLM $M$ (i.e., without $U$ and $\Delta W_{i=1}^{|W|}$), and pool the output representations $\{\mathbf{x}_{t_1}, ..., \mathbf{x}_{|T_M|}\}$ of tokens $\{t_{1}, ..., t_{|T_M|}\}$ to a sequence embedding $\phi\left(\{x_t\}_{t=1}^{T_M}\right) = \mathbf{\bar{x}}^{S}$. In the subsequent step, we first re-embed the sequence $S$ with the MT encoder $E$. We then input the MT encoder output embeddings into the LLM $M$, now including $U$ and $\Delta W_{i=1}^{L}$, and again pool the resulting output representations  $\{\mathbf{z}_{1}, ..., \mathbf{z}_{|T_E|}\}$ of tokens $\{t_{1}, ..., t_{|T_E|}\}$ to a sequence embedding $\phi\left(\{\mathbf{z}_t\}_{t=1}^{T_E}\right) = \mathbf{\bar{z}}^{S}$. The appropriate pooling function $\phi$ depends on the prior training regime of the LLM. Common choices include \texttt{BOS}-pooling $\phi\left(\{\mathbf{x}_t\}_{t=1}^{T}\right) = \mathbf{x}_1$ or mean-pooling $\phi\left(\{\mathbf{x}_t\}_{t=1}^{T}\right) = \frac{1}{T} \sum_{i=1}^{T} \mathbf{x}_i$ for encoders, as well as \texttt{EOS}-pooling $\phi\left(\{\mathbf{x}_t\}_{t=1}^{T}\right) = \mathbf{x}_T$ for decoders.
We train the parameters $\Theta$ (i.e, $U$ and $\Delta W_{i=1}^{L}$) to minimize the mean-squared error $\text{\texttt{MSE}}( \mathbf{\bar{x}}^{S}, \mathbf{\bar{z}}^{S})$.

\rparagraph{Stage 2: Task-Specific Distillation} The second stage is designed to bridge the remaining misalignment between the MT encoder and LLM in our fused MT-LLM $E \times M$ in task-specific fine-tuning. We hypothesize that optimizing MT-LLM's representation alignment on general-purpose data is less sample-efficient than task-specific alignment. Figure \ref{fig:nllb-llm2vec-task-distillation} depicts the task-specific distillation process. 

We first fine-tune the base LLM with a classification head $H \in \mathbb{R}^{d_m \times |C|}$ on the labeled task training data. Task fine-tuning reduces the complexity of the LLM's output representations, reducing them to encoding only task-specific features: this, in turn, facilitates task-specific representational alignment for our MT-LLM. Similar to the previous, adaptation stage, we then again fine-tune only the parameters of the LoRA adapters $\Theta$: we continue training the LoRA adapters obtained in Stage 1 (i.e., task-agnostic adaptation).\footnote{In our preliminary experiments, merging the LoRAs of the adaptation phase with new, fine-tuning LoRA adapters led to numerical instabilities of weights due to quantization, which severely degraded final task performance.} We again minimize the mean-squared error $\text{\texttt{MSE}}(\mathbf{\bar{x}}^{S}, \mathbf{\bar{z}}^{S})$. During inference, we classify instances with the task head $H$ as trained in the initial LLM task fine-tuning.
This way we improve the model's ability to generalize on the task, as the MT-LLM is trained to match the output of the knowledge-rich task-specific representations of the fine-tuned LLM.  

Both alignment steps together ensure that the latent translations from the MT encoder seamlessly integrate as input representations into the LLM backbone. And this integration extends the access to the knowledge embedded in the LLM to all languages supported by the MT model.

\label{sec:methodology}

\section{Experimental Setup}
\label{sec:experimental-setup}
\subsection{Tasks and Languages}

We evaluate on two established classification tasks and one multiple-choice machine reading comprehension (MRC) task, which all require nuanced NLU capabilities. For each task-dataset combination, we evaluate on all languages supported by the selected underlying MT model.\footnote{Appendix \ref{app:full-results} lists the full details.}

\vspace{0.5mm}
\noindent \textit{Natural Language Inference (NLI)}.
We evaluate on XNLI~\citep{conneau-etal-2018-xnli}, AmericasNLI (AmNLI)~\citep{ebrahimi-etal-2022-americasnli}, and the NLI data of Kardeş-NLU~\citep{senel-etal-2024-kardes}. 
We fine-tune models on the training portion of MNLI~\citep{williams-etal-2018-broad}. We feed the mean-pooled token representations of the jointly embedded hypothesis-premise sentence-pair into the classifier.

\vspace{0.5mm}
\noindent \textit{Sentiment Classification} is
evaluated on NusaX \cite{winata-etal-2023-nusax}, which encompasses 10 Indonesian languages.\footnote{In our experiments on Buginese, as the only outlier, \zsxlt performance progressively deteriorated with more distillation (cf. Table \ref{tab:nusax-by-step}). This is in line with unusual behavior for Buginese reported in other work using NLLB \cite{ebing2023translate}. We thus exclude Buginese results from the main discussion.} We use the English training, and validation splits with 500 and 100 instances, respectively, as source-language data. We feed the mean-pooled token embeddings of the input text into the classifier.

\vspace{0.5mm}
\noindent \textit{Multiple-Choice MRC}.
Belebele is a multiple-choice MRC benchmark encompassing 122 typologically diverse language variants~\cite{bandarkar2023belebele}. We train models on the English training data provided by~\citet{bandarkar2023belebele}.
We jointly embed the paragraph, question, and answers. For each choice $c_i \in C$, we then average the token embeddings and regress the resulting representation via head $H^{d_M \times 1}$ to a logit $\mathbf{l_{c_i}}$. We then minimize the cross-entropy between the concatenated choice logits $\{\mathbf{l_{c_i}}\}_{i=1}^{|C|}$ and the true label.

\subsection{Cross-Lingual Transfer Setups}
We evaluate \xlt abilities of LLMs in two standard setups, \zsxlt and \ttest. Both paradigms enable \xlt without requiring further annotation or prolonged training for any target language.
We do not evaluate \ttrain or involved strategies based on back-translations of source-language training data \cite{artetxe-etal-2023-revisiting,ebing2023translate} as they require computationally intensive task-specific fine-tuning, independently for \textit{each} target language; these variants also require sufficient target language `understanding' ability from the LLM, which is not there for low-resource languages.\footnote{Unlike the standalone LLM, our MT-LLM supports \ttrain to any target language supported by the MT encoder.}

\rparagraph{\zsxlt} In \zsxlt, the model is first trained on source-language training data. Since the model is multilingual, \xlt is inherently supported: we simply run inference on target-language instances. Since LLMs are not sufficiently pretrained multilingually, we align them with an MT encoder with our self-distillation procedure (cf. Stage 1 in \S\ref{sec:methodology}).

\rparagraph{\ttest}
In \ttest, the model is initially trained on labeled source-language instances. During inference, the target-language instances are translated to the source language prior to prediction with a dedicated MT model. This enables \xlt with monolingual (L)LM backbones.

\subsection{Models and Training Setup}

\rparagraph{Translation Models}
We use the NLLB 600M parameter model as our primary MT encoder backbone~\cite{nllbteam2022language} for MT-LLM distillation. For \ttest, we translate validation and test datasets with both the 600M NLLB model as well as with the larger, 3.3B parameter variant. We use greedy decoding as \citet{ebing2023translate} showed that more sophisticated decoding strategies yield no downstream improvements in \xlt.

\rparagraph{LLMs}
We base our experiments on the Llama 3-8B variant~\cite{llama3modelcard} that underwent the `\llmvec process' \cite{behnamghader2024llm2vec}. \llmvec is a recipe that converts decoder-only LLMs into powerful sequence encoders by (i) enabling bidirectional attention and continuing training on both (ii) self-supervised masked next-token prediction, and (iii) SimCSE \cite{gao-etal-2021-simcse}.\footnote{We refer the reader to the original \llmvec work for further technical details.} We refer to the model that fuses the NLLB 600M encoder with \llmvec as NLLB-\llmvec. We then adapt to downstream tasks by performing either direct fine-tuning on labeled instances or task-specific self-distillation (cf. Stage 2 in \S\ref{sec:methodology}).

\rparagraph{Training Details}
We train all models using LoRAs with rank $r{=}16$, alpha $\alpha{=}32$, and LoRA dropout of $0.05$ inserted into all linear layers. We further train models with the 8-bit AdamW~\citep{loshchilov2018decoupled,DBLP:journals/corr/abs-2110-02861}, 4-bit QLoRA-style quantization \cite{dettmers2023qlora}, weight decay of $0.01$, and with 10\% linear warm-up and then linear decay. Experimental results are averaged across three random seeds.\footnote{For NusaX, we repeat experiments with 5 random seeds due to the smaller dataset size.}

\rparagraph{\textit{Stage 1: Setup}} We train for $10$K steps on the 10B tokens subsampled from the FineWeb corpus~\cite{penedo2024fineweb}. While our approach supports simultaneous adaptation on all languages supported by both the MT model and the LLM, we adapt the LLM only on English text.\footnote{By unlocking \zsxlt via self-supervised distillation on English text alone, we demonstrate that any monolingual language model can be equipped with an MT encoder.} We set the effective batch size to 256. Learning rate is $2e^{-4}$.

\rparagraph{\textit{Baselines and Stage 2: Setup}} We set the learning rate to $1e^{-4}$ for downstream task experiments. We fine-tune models with an effective batch size of 32, for 3 epochs on NLI, for 5 epochs on Belebele, and for 20 epochs on NusaX. We validate models at every 10\% of total training steps. We validate and test on all languages that are supported by our MT model. We start task-specific self-distillation from model snapshots that performed best on source-language validation instances.

\section{Results and Discussion}
\label{sec:results}
  \begin{table*}[t!]
  \begin{adjustbox}{width=\textwidth,center}

     \begin{tabular}{@{\extracolsep{\fill}} l!{\vrule width \arrayrulewidth}cc!{\vrule width \arrayrulewidth}cc!{\vrule width \arrayrulewidth}cc!{\vrule width \arrayrulewidth}cc!{\vrule width \arrayrulewidth}c}
        \toprule
        \multicolumn{1}{c}{} & \multicolumn{2}{c}{\textbf{\textsc{XNLI}}} & \multicolumn{2}{c}{\textbf{\textsc{AmNLI}}} & \multicolumn{2}{c}{\textbf{\textsc{Kardeş-NLU}}} & \multicolumn{2}{c}{\textbf{\textsc{NusaX}}} & \textbf{\textsc{Belebele}} \\
        \cmidrule(lr){2-3} \cmidrule(lr){4-5} \cmidrule(lr){6-7} \cmidrule(lr){8-9} \cmidrule(lr){10-10}
        & \sdev & 	\tdev & \sdev & 	\tdev & \sdev & 	\tdev & \sdev & 	\tdev & \sdev \\
        \hline
        \rowcolor{gray!10} \multicolumn{10}{l}{\rule{0pt}{2.5ex} \textit{Zero-Shot Cross-Lingual Transfer: Fine-tune multilingual model on English training set}}  \\ 
        \hline
        \llmvec         & $68.9_{\pm2.0}$ & $71.1_{\pm2.4}$ & $40.9_{\pm2.0}$ & $43.2_{\pm1.6}$ & $46.7_{\pm1.7}$ & $51.1_{\pm13.2}$ & $54.5_{\pm13.7}$ & $58.9_{\pm10.9}$ & $48.2_{\pm3.2}$ \\
        NLLB-Encoder    & $71.6_{\pm0.2}$ & $71.8_{\pm0.3}$ & $55.3_{\pm0.6}$ & $56.4_{\pm0.3}$ & $74.9_{\pm0.5}$ & $75.2_{\pm0.6}$ & $80.7_{\pm0.2}$ & $81.7_{\pm0.1}$ & $30.4_{\pm0.4}$ \\
        NLLB-\llmvec S1+FT & $80.0_{\pm0.9}$ & $80.4_{\pm0.4}$ & $63.0_{\pm1.9}$ & $64.3_{\pm1.2}$ & $81.5_{\pm1.3}$ & $81.3_{\pm0.4}$ & $72.7_{\pm4.4}$ & $77.5_{\pm2.4}$ & $60.2_{\pm0.5}$ \\
        NLLB-\llmvec S1+S2 & $\mathbf{81.4_{\pm0.6}}$ & $\mathbf{81.7_{\pm0.5}}$ & $\mathbf{64.0_{\pm0.3}}$ & $\mathbf{64.6_{\pm0.7}}$ & $\mathbf{82.2_{\pm0.5}}$ & $\mathbf{82.1_{\pm0.5}}$ & $82.1_{\pm2.6}$ & $\mathbf{82.6_{\pm2.4}}$ & $62.6_{\pm0.5}$ \\
        \hline
        \rowcolor{gray!10} \multicolumn{10}{l}{\rule{0pt}{2.5ex} \textit{Translate-Test: Translate test data to English}} \\ 
        \hline
        \llmvec NLLB-600M & $78.7_{\pm0.7}$ & $78.6_{\pm0.9}$ & $52.0_{\pm0.7}$ & $52.7_{\pm0.6}$ & $78.8_{\pm0.8}$ & $78.4_{\pm1.0}$ & $78.3_{\pm0.9}$ & $78.8_{\pm1.2}$ & $60.7_{\pm0.7}$ \\
        \llmvec NLLB-3B  & $80.2_{\pm0.6}$ & $80.2_{\pm0.8}$ & $50.9_{\pm0.4}$ & $51.2_{\pm1.7}$ & $79.9_{\pm0.9}$ & $79.9_{\pm1.0}$ & $\mathbf{82.4_{\pm0.6}}$ & $\mathbf{82.6_{\pm0.5}}$ & $\mathbf{64.2_{\pm0.7}}$ \\
        \bottomrule
    \end{tabular}
\end{adjustbox}
\caption{\textbf{\zsxlt vs. \ttest.} We benchmark models on \zsxlt against \ttest on non-English NLU test sets (cf. \S\ref{sec:experimental-setup}). S1 and S2 refer to self-supervised and task-specific stages of aligning NLLB with \llmvec (cf. \S\ref{sec:methodology}). FT denotes supervised fine-tuning. Reported performance is averaged over three seeds on model checkpoints that maximize performance on source-language (\sdev) and per target-language (\tdev) validation splits. Subscripts denote std. deviation. Metrics: accuracy for NLI and Belebele, macro-F1 for NusaX. Best model per column is in \textbf{bold}.}
\label{tab:main-results-zsxlt}
\end{table*}
Table \ref{tab:main-results-zsxlt} summarizes the results for each task, dataset, and model configuration. We then analyze the results per each of these dimensions. 
\smallskip

\rparagraph{\zsxlt} Following prior work~\cite{schmidt-etal-2023-free}, we report final \xlt test performance for model checkpoints that maximize performance on the source-language (\sdev) and target-language (\tdev) validation splits, respectively, in order to estimate the bounds of both expected (\sdev) and ideal \zsxlt performance (\tdev). \tdev also absorbs fluctuation in transfer performance stemming from sub-optimal hyperparameters \cite{keung-etal-2020-dont, schmidt-etal-2023-free}. 
\smallskip

\noindent \textit{LLM2Vec}.
Despite its strong English performance (as demonstrated in Table \ref{tab:main-results-eng}), the English-centric \texttt{\llmvec} model based on Llama 3 8B underperforms all other models in \zsxlt. For instance, the considerably smaller NLLB-Encoder (413M parameters) alone outperforms \llmvec on both the NLI and NusaX tasks. This confirms that LLMs generally underperform in NLU tasks for languages other than English. Notably, \ttest significantly improves upon the \zsxlt performance of \llmvec, especially on datasets that predominantly feature lower-resource languages (AmNLI, Kardeş-NLU). Expectedly, and consistent with findings from related work \cite{ansell-etal-2023-unifying}, the larger MT model (NLLB 3B) improves the \ttest performance on all tasks.

\noindent \textit{NLLB-LLM2Vec}.
The NLLB encoder alone shows strong \zsxlt performance on sentiment classification (NusaX) but performs worse on NLI and degrades on MRC (Belebele), as the more intricate NLU task. This suggests that MT encoders indeed lack language understanding abilities and knowledge typically acquired with LM objectives on large-scale corpora. 
Our integrated NLLB-\llmvec variants substantially outperform both LLM2Vec and NLLB-Encoder on all NLU tasks, with performance gains on Belebele of 12\% and 30\%, respectively. 
Specifically, fine-tuning the NLLB-\llmvec adapted only in the task-agnostic manner (S1+FT) is already competitive with the more computationally involved \ttest. 
Our secondary task-specific distillation, i.e., NLLB-\llmvec S1+S2, further substantially and consistently improves the performance compared to direct fine-tuning (NLLB-\llmvec S1+FT). The gains are particularly prominent on NusaX ($+9.4\%$), which has the smallest training set. Our full NLLB-\llmvec S1+S2 consistently beats \ttest based on the same NLLB 600M model by sizable margins ($3-11\%$). What is more, NLLB-\llmvec S1+S2 frequently performs on par or better than \ttest that uses higher-quality translations from the larger NLLB 3B: MT improvements do propagate to \zsxlt because of favorable model selection on \tdev. These results show that NLLB-\llmvec (S1+S2) boosts \zsxlt by allowing the NLU abilities of the LLM to propagate to many languages via the high-quality multilingual representation space of the NLLB encoder.
\smallskip

In sum, our alignment procedure effectively merges NLLB into \llmvec to enable \zsxlt that both outperforms \ttest and reduces inference cost by avoiding decoding in the MT model. Consistent improvements of \ttest with NLLB 3B over NLLB 600M suggests that further \zsxlt gains can be seized by integrating larger MT models into \llmvec.  Moreover, since NLLB-\llmvec is inherently multilingual, NLLB-\llmvec can further benefit from training on translated training data (i.e., \ttrain): NLLB-\llmvec is poised to robustly encode noisily translated training data, as it was trained both on MT and denoising autoencoding \cite{nllbteam2022language}. 

\begin{table}[t!]
  \begin{adjustbox}{width=\columnwidth,center}
    \begin{tabular}{@{\extracolsep{\fill}} l!{\vrule width \arrayrulewidth}c!{\vrule width \arrayrulewidth}c!{\vrule width \arrayrulewidth}c}

        \toprule
        \multicolumn{1}{c}{} & \multicolumn{1}{c}{\textsc{\textbf{XNLI}}} & \multicolumn{1}{c}{\textsc{\textbf{NusaX}}} & \multicolumn{1}{c}{\textsc{\textbf{Belebele}}} \\
        \midrule
        \llmvec & $\mathbf{92.5_{\pm0.3}}$ & $91.3_{\pm0.5}$ & $\mathbf{94.0_{\pm0.4}}$ \\ \hline
        NLLB-Encoder    & $80.4_{\pm0.2}$ & $86.9_{\pm0.2}$ & $33.6_{\pm0.1}$ \\
        NLLB-\llmvec S1+FT & $90.0_{\pm0.7}$ & $90.8_{\pm0.6}$ & $91.0_{\pm1.0}$ \\
        NLLB-\llmvec S1+S2 & $91.4_{\pm0.2}$ & $\mathbf{92.2_{\pm0.5}}$ & $92.4_{\pm0.7}$ \\
        \bottomrule
    \end{tabular}
  \end{adjustbox}
  \caption{\textbf{English performance.} We benchmark \llmvec, the NLLB encoder, and our fused NLLB-\llmvec on English test sets of various NLU benchmarks (cf. \S\ref{sec:experimental-setup}). See Table \ref{tab:main-results-zsxlt} for further details.}
  \label{tab:main-results-eng}
\end{table}

\rparagraph{English} Table \ref{tab:main-results-eng} shows the in-language (i.e., no \xlt) English performance by task for our models.
\smallskip

\noindent \textit{LLM2Vec}. Pre-trained on English-dominated web-scale corpora,  \llmvec demonstrates strong performance on all tasks. The comparison of in-language performance in Table \ref{tab:main-results-eng} with \zsxlt performance in Table \ref{tab:main-results-zsxlt} shows the scale of performance drop for \llmvec in \xlt. This means that LLMs require either extensive multilingual pre-training or post-hoc language adaptations for effective \xlt.
\smallskip

\noindent \textit{NLLB-\llmvec}. The results for NLLB variants, the NLLB encoder alone and our NLLB-\llmvec, provide more context for the \zsxlt results from Table \ref{tab:main-results-zsxlt}. 
We observe that \zsxlt performance of NLLB variants is correlated with their in-language English performance. While the NLLB Encoder performs fairly on NusaX (and to some extent also on NLI), it lacks language understanding abilities to 
that match more complex NLU tasks like Belebele. 
Our NLLB-\llmvec variants, on the other hand, successfully exploit the knowledge of Llama to materially increase English performance over the NLLB Encoder ($+58\%$ on Belebele). Our task-agnostic NLLB-\llmvec alignment (S1+FT) still lags somewhat behind \llmvec after fine-tuning on labeled task data.
We manage to narrow this gap for Belebele and NLI with task-specific distillation (S1+S2) (cf. \S\ref{sec:methodology}) and even surpass the English performance of the LLM for NusaX. 
This suggests that the task-specific distillation guides NLLB-\llmvec to better leverage the knowledge embedded in the weights of \llmvec, and shape it specifically for the task. 
\smallskip

The results indicate that compositional alignment on the word- or span-level, as introduced in the task-distillation on the Belebele dataset, further improves representational alignment in the MT-LLM. As evident from the comparison of English results in Table \ref{tab:main-results-eng} and \xlt results in Table \ref{tab:main-results-zsxlt}, better global (i.e., task-agnostic) alignment, in turn, directly transfers to closing the `English knowledge gap', i.e., to better \zsxlt performance.

\subsection{Further Analyses and Discussion}

\begin{figure}
    \centering
    \includegraphics[scale=1.11]{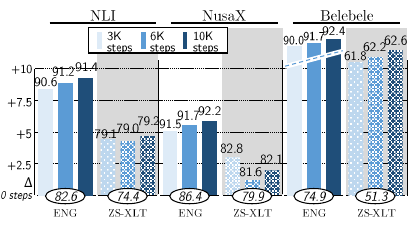}
    \caption{\textbf{Impact of Adaptation.} We evaluate our 2-step alignment procedure by $\{0,3,6,10\}$K general adaptation steps (Stage 1) (cf. \S\ref{sec:methodology}) on English (\textsc{ENG}) and non-English (\zsxlt) test portions of various NLU benchmarks. Model selection on \sdev.}
    \label{fig:nllb-llm2vec-sample-efficiency}
\end{figure}

\rparagraph{Importance of Adaptation} 
Figure \ref{fig:nllb-llm2vec-sample-efficiency} shows both English and \zsxlt performance by task for task-specific self-distillation, NLLB-\llmvec (S1+S2), after $K \in \{0, 3, 6, 10\}$ steps of task-agnostic alignment (see \S\ref{sec:methodology}), respectively.
The figure points to the importance of task-agnostic adaptation both for English and \zsxlt performance. The results furthermore highlight that Stage 1 of our alignment is sample-efficient: the largest relative \zsxlt gains are obtained after only 3K training steps (e.g., $+10.5\%$ for Belebele) and then marginalize with further training. We observe the same trends for the English performance (e.g., $+15\%$ on Belebele from 3K steps of alignment). These results show that we can effectively tie LLMs and MT encoders into a unified multilingual MT-LLM at computational cost that is negligible w.r.t. both LLM and MT (pre-)training.

On NusaX, we observe that while prolonged task-agnostic adaptation benefits the in-language English performance, it does not improve \zsxlt results. 
The explanation, we believe, is in the simplicity of the task: \zsxlt performance on NusaX saturates quickly because the NLLB encoder already solves the task well (see\, Table \ref{tab:main-results-zsxlt}) and thus requires little additional knowledge from \llmvec, to which it gets access through the alignment.

The results indicate that the gap in English performance between \llmvec and NLLB-\llmvec (cf. Table \ref{tab:main-results-eng}) can eventually be closed with longer alignment. We also hypothesize that explicit token or span alignment objectives would improve the generalization: this would be facilitated by the significant overlap between the vocabularies of Llama 3 and NLLB tokenizers.

  \begin{table}[t!]
  \begin{adjustbox}{width=\columnwidth,center}

    \begin{tabular}{@{\extracolsep{\fill}} l!{\vrule width \arrayrulewidth}c!{\vrule width \arrayrulewidth}cc!{\vrule width \arrayrulewidth}cc!{\vrule width \arrayrulewidth}cc}
        \toprule
        \multicolumn{1}{l}{} & \multicolumn{3}{c}{\textbf{\textsc{XNLI}}} & \multicolumn{2}{c}{\textbf{\textsc{AmNLI}}} & \multicolumn{2}{c}{\textbf{\textsc{KNLU}}}\\
        \cmidrule(lr){2-4} \cmidrule(lr){5-6} \cmidrule(lr){7-8} 
        \textit{GPT-2} & \textsc{eng} & \textsc{s-d} & 	\textsc{t-d} & \textsc{s-d} & 	\textsc{t-d} & \textsc{s-d} & 	\textsc{t-d}  \\
        \hline
        \rowcolor{gray!10} \multicolumn{8}{l}{\rule{0pt}{2.5ex} \textit{Zero-Shot Cross-Lingual Transfer}}  \\ 
        \hline
        \textcolor[gray]{0.4}{NLLB Enc.} & \textcolor[gray]{0.4}{$80.4$} & \textcolor[gray]{0.4}{$71.6$} & \textcolor[gray]{0.4}{$71.8$} & \textcolor[gray]{0.4}{$55.3$} & \textcolor[gray]{0.4}{$56.4$} & \textcolor[gray]{0.4}{$74.9$} & \textcolor[gray]{0.4}{$75.2$} \\

        NLLB FT & $82.2$ & $74.7$ & $74.8$ & $\mathbf{62.4}$ & $\mathbf{63.2}$ & $\mathbf{76.2}$ & $\mathbf{76.5}$ \\
        \hline
        \rowcolor{gray!10} \multicolumn{8}{l}{\rule{0pt}{2.5ex} \textit{Translate-Test}} \\ 
        \hline
        NLLB-600M & $\mathbf{85.0}$ & $73.6$ & $74.0$ & $54.1$ & $55.2$ & $74.7$ & $75.3$ \\
        NLLB-3B  & $\mathbf{85.0}$ & $\mathbf{75.1}$ & $\mathbf{75.4}$ & $52.6$ & $54.5$ & $75.4$ & $75.9$ \\
        \bottomrule
    \end{tabular}
\end{adjustbox}
\caption{\textbf{Adaptation on GPT-2.} We perform adaptation (stage 1) with GPT-2 and benchmark NLLB-GPT-2 against GPT-2 in English and non-English test portions of NLI benchmarks.  We repeat NLLB Encoder results in gray as a reference. See Table \ref{tab:main-results-zsxlt} for further details.}
\label{tab:gpt2-results}
\end{table}

\rparagraph{Fusing Decoder Models with MT Encoders} 
We additionally test the integration of MT encoders into a decoder LM: we align the NLLB 600M encoder to the GPT-2 medium (354M parameters) and evaluate on NLI.\footnote{We exclude NusaX and Belebele for this ablation: (1) the NLLB encoder performs better than \llmvec in the \ttest setting on NusaX; (2) For Belebele, the limited context length of GPT-2 hinders a fair comparison.} Due to the absence of the \texttt{EOS} token in the pretraining of GPT-2, we perform task-agnostic self-distillation (Stage 1, on the FineWeb corpus) using mean-pooled token representations (cf. \S\ref{sec:methodology}).
Subsequently, we fine-tune the NLLB-GPT-2 directly on MNLI, feeding the \texttt{EOS}-pooled representations into the classifier.\footnote{We omit task-specific distillation because it performed slightly worse on English in XNLI compared to directly fine-tuning of NLLB-GPT-2 with larger learning rates.} We increase the learning rate to $3e^{{-}4}$ and leave other hyperparameters unchanged. Like in our main experiments, we compare NLLB-GPT-2 against \ttest with NLLB-600M and NLLB-3.3B, respectively.

The results in Table \ref{tab:gpt2-results} show that NLLB-GPT-2 successfully taps into GPT-2's `knowledge' to outperform both the standalone NLLB encoder and the \textit{fair} \ttest baseline using the same NLLB-600M model in \zsxlt across all datasets. Moreover, NLLB-GPT-2 even surpasses \ttest on GPT-2 with the larger NLLB-3.3B model on \textsc{AmNLI} and \textsc{Kardeş-NLU}. These results hold despite the suboptimal alignment, as indicated by the notable gap in performance to the fine-tuned GPT-2 on the English test portion of XNLI. The discrepancy likely stems from challenges in converting encoders into decoders, as observed in prior work \cite{pmlr-v162-wang22u}. We believe that prolonged adaptation and explicit token-level alignment objectives would further improve both sample-efficiency and quality of alignment, reducing the `knowledge' gap.

\section{Conclusion}
LLMs quickly emerged as the catch-all solution to NLU in English. However, LLMs still cannot extend their NLU abilities to languages typologically distant from English or virtually unseen at pretraining. In this work, we propose a novel approach to fuse MT encoders with LLM backbones via self-distillation to compile a massively multilingual MT-LLM. The MT-LLM not only strongly improves \zsxlt performance over \ttest but also removes the overhead of MT decoding at inference. We demonstrate that our distillation procedure is highly efficient and requires only a few thousand steps to convert LLMs into multilingual MT-LLMs, enabling NLU in all languages supported by the MT encoder. We further show that our MT-LLM alignment benefits both encoder and decoder LLMs. 
In future work we will seek to (1) further improve generalization of MT-LLM by incorporating token-level alignment objectives and (2) extend the MT-LLM to support further languages by post-hoc adaptation of  the MT encoder. 

\section{Limitations}

Our experimental results are based on using Llama 3 and GPT-2 as the LLM backbones and NLLB-600M as the MT encoder in our MT-LLM approach. Expanding our experimental setup to include a wider range of MT encoders and additional LLM backbones would not only validate its applicability across various model families and architectures but also enrich our findings.
The Llama 3 backbone of \llmvec underwent instruction tuning. This means that \llmvec might have seen labelled data for tasks we experiment on in our work. We strongly believe this does not constitute an issue to evaluate \textit{cross-lingual} transfer of our model configurations. If there was serious leakage of labelled instances, \ttest variants should benefit more strongly as the data is presented in the language Llama was trained on.
Our method would gain further support, if our approach extended to generative language modelling. However, sequence-level alignment objectives do not sufficiently align the MT and LLM backbones. The MT and LLM backbones therefore would require either matching or largely overlapping vocabularies to appropriately learn how to fuse the models on the token level (cf. \S\ref{sec:methodology}).
Another non-negligible consideration in our evaluations that our limited compute budget does not allow for is hyperparameter tuning. We nevertheless believe our main evaluations put model variants on equal footing and hence reliably measure expected \zsxlt. We further counteract this issue in two ways. First, prior work shows that LoRAs are generally more robust to varying hyperparameters. Second, we report transfer performance both on when selecting models on source-language and per target-language validation. The latter remedies oscillation in \zsxlt performance \cite{keung-etal-2020-dont,schmidt-etal-2023-free}.

\bibliography{anthology,custom}

\appendix

\section{Appendix}
\label{sec:appendix}
\subsection{Reproducibility details}

\sparagraph{Compute Requirements} We perform general-purpose adaptation (i.e., stage 1, cf. \S\ref{sec:methodology}) on 8 A100 80GB, which requires about 22 hours of runtime. All downstream experiments were executed on A100 40GB. Downstream fine-tuning and distillation required for each one of three seeds ca. 20 hours of runtime for NLI, ca. 30 hours of runtime of Belebele, and ca. 20 minutes of runtime per NusaX. We execute these experiments for \llmvec fine-tuning, NLLB-\llmvec S1+FT, and NLLB-\llmvec S1+S2 (cf. \S\ref{sec:methodology}). The compute required for downstream fine-tuning therefore sums roughly to 450 GPU hours.
Subsequent evaluations required, per each of ten evaluated checkpoints, about 3 hours on XNLI, AmNLI, and Kardeş-NLU combined, 5 hours on Belebele, and 10 minutes on NusaX. We estimate that inference therefore requires 725 hours of GPU runtime.
In conclusion, our experiments in total required between 50 to 60 days of A100 runtime.

\sparagraph{Code} We will make the code publicly available at \url{https://github.com/fdschmidt93/trident-mt-llm}. For the preprint under review, the code is available as an attachment to the submission.

\sparagraph{Translations} The translated validation and test splits for all datasets (cf. \S\ref{sec:methodology}) are available via the Github repository.

\sparagraph{Additional Details on Experimental Setup}

Please refer to Table \ref{app:datasets-tab} for details on the number of languages and instances by dataset and split. In what follows, we outline how we accessed the datasets.

\noindent\textit{MNLI.}  We access the training portion of the MNLI dataset via Hugging Face at \url{https://huggingface.co/datasets/nyu-mll/glue}.

\noindent\textit{XNLI.}  We access the training portion of the MNLI dataset via Hugging Face at \url{https://huggingface.co/datasets/nyu-mll/glue}.

\noindent\textit{AmNLI.}  We access the training portion of the MNLI dataset via Hugging Face at \url{https://huggingface.co/datasets/nala-cub/americas_nli}.

\noindent\textit{Kardeş-NLU.}  Our code includes a script to access the dataset via the Hugging Face \texttt{datasets} framework. The original dataset is available at: \url{https://github.com/lksenel/Kardes-NLU}.

\noindent\textit{NusaX.}  We access the dataset via Hugging Face at \url{https://huggingface.co/datasets/indonlp/NusaX-senti}.

\noindent\textit{Belebele.} The training dataset is available at \url{https://github.com/facebookresearch/belebele}. We access the dataset via Hugging Face at \url{https://huggingface.co/datasets/facebook/belebele}.

\noindent\textit{LLM2Vec.}  We use \llmvec that has been trained without supervision which is available on the Hugging Face hub: \href{https://huggingface.co/McGill-NLP/LLM2Vec-Meta-Llama-3-8B-Instruct-mntp-unsup-simcse}{here}.

\noindent\textit{NLLB.}  The distilled 600M parameters variant of NLLB is available at \url{https://huggingface.co/facebook/nllb-200-distilled-600M}. NLLB 3B can be accessed via \url{https://huggingface.co/facebook/nllb-200-3.3B}.

\begin{figure}
\subsection{NLLB vs. XLM-R on XNLI}
\label{subsec:nllb-vs-xlmr}
     \centering
     \includegraphics[scale=1.9]{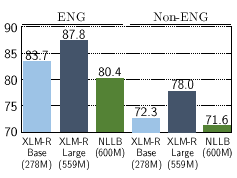}
     \vspace{-1.0em} 
     \caption{Performance on XNLI English and non-English test portions when selecting the model on English validation instances (cf. \S\ref{sec:experimental-setup}) for XLM-R-Base, XLM-R-Large, and the NLLB-Encoder \cite{conneau-etal-2020-unsupervised, nllbteam2022language}}.
     \label{fig:xnli-nllb-xlmr}
 \vspace{-0.5em}
 \end{figure}

\begin{table*}[t!]
\subsection{Datasets}
\label{app:datasets-tab}
\begin{center}
    \begin{tabular}{@{\extracolsep{\fill}} l!{\vrule width \arrayrulewidth}c!{\vrule width \arrayrulewidth}c!{\vrule width \arrayrulewidth}c!{\vrule width \arrayrulewidth}c!{\vrule width \arrayrulewidth}c!{\vrule width \arrayrulewidth}c!{\vrule width \arrayrulewidth}c}
        \toprule
         & \textbf{\textsc{MNLI}} & \textbf{\textsc{XNLI}} & \textbf{\textsc{AmNLI}} & \textbf{\textsc{Kardeş-NLU}} & \textbf{\textsc{NusaX}} & \textbf{\textsc{Belebele}} \\
        \hline
        No. of Languages & 1 & 15 & 3 & 4 & 9 & 117 \\ \hline
        Training   & $392,702$ & $--$    & $--$ & $--$     & $500$ & $67,541$ \\
        Validation & $--$      & $2,490$ & $743$ & $1,000$ & $100$ & $3,773$ \\
        Test       & $--$      & $5,010$ & $750$ & $2,000$ & $400$ & $900$ \\
        \bottomrule
    \end{tabular}
\caption{Number of instances by split by language per dataset. For NLI, we train on the training split of MNLI~\cite{williams-etal-2018-broad}. Number of languages denominates the number of languages supported by NLLB. For Belebele, we construct training and validation datasets with the scripts provided by ~\cite{bandarkar2023belebele} (cf. Appendix \ref{sec:appendix}).}
\end{center}
\label{tab:metrics}
\end{table*}

\begin{table*}[t!]
\subsection{Full results}
\label{app:full-results}
\rparagraph{Main Results}\\

\begin{adjustbox}{width=\textwidth,center}
    \begin{tabular}{@{\extracolsep{\fill}} l!{\vrule width \arrayrulewidth}c    !{\vrule width \arrayrulewidth}!{\vrule width \arrayrulewidth}    c!{\vrule width \arrayrulewidth}c!{\vrule width \arrayrulewidth}c!{\vrule width \arrayrulewidth}c!{\vrule width \arrayrulewidth}c!{\vrule width \arrayrulewidth}c!{\vrule width \arrayrulewidth}c!{\vrule width \arrayrulewidth}c!{\vrule width \arrayrulewidth}c!{\vrule width \arrayrulewidth}c!{\vrule width \arrayrulewidth}c!{\vrule width \arrayrulewidth}c!{\vrule width \arrayrulewidth}c!{\vrule width \arrayrulewidth}c!{\vrule width \arrayrulewidth}c}
        \toprule
         & \textbf{\textsc{EN}} & \textbf{\textsc{AR}} & \textbf{\textsc{BG}} & \textbf{\textsc{DE}} & \textbf{\textsc{EL}} & \textbf{\textsc{ES}} & \textbf{\textsc{FR}} & \textbf{\textsc{HI}} & \textbf{\textsc{RU}} & \textbf{\textsc{SW}} & \textbf{\textsc{TH}} & \textbf{\textsc{TR}} & \textbf{\textsc{UR}} & \textbf{\textsc{VI}} & \textbf{\textsc{ZH}}  & \textbf{\textsc{AVG}} \\
        \hline
        \rowcolor{gray!10} \multicolumn{17}{l}{\rule{0pt}{2.5ex} \textit{Zero-Shot Cross-Lingual Transfer: Fine-tune multilingual model on English training set}}  \\ 
        \hline
        \llmvec            & $92.5_{\pm0.3}$ & $63.6_{\pm5.1}$ & $68.5_{\pm3.8}$ & $79.8_{\pm1.3}$ & $73.0_{\pm1.2}$ & $84.5_{\pm0.8}$ & $83.0_{\pm0.9}$ & $55.8_{\pm8.3}$ & $78.0_{\pm1.4}$ & $43.0_{\pm3.6}$ & $66.6_{\pm0.9}$ & $67.3_{\pm2.4}$ & $45.2_{\pm4.1}$ & $78.0_{\pm1.3}$ & $78.7_{\pm0.9}$ & $68.9_{\pm2.0}$ \\
        NLLB-Encoder       & $80.4_{\pm0.2}$ & $71.3_{\pm0.8}$ & $75.2_{\pm0.1}$ & $74.9_{\pm0.4}$ & $74.5_{\pm0.0}$ & $75.5_{\pm0.4}$ & $75.3_{\pm0.4}$ & $68.6_{\pm0.8}$ & $72.7_{\pm0.2}$ & $69.5_{\pm0.3}$ & $65.7_{\pm1.0}$ & $72.6_{\pm0.4}$ & $65.1_{\pm0.7}$ & $72.8_{\pm0.2}$ & $69.1_{\pm0.6}$ & $71.6_{\pm0.2}$ \\
        NLLB-\llmvec S1+FT & $90.7_{\pm0.4}$ & $80.9_{\pm1.2}$ & $85.6_{\pm0.9}$ & $84.9_{\pm0.8}$ & $73.8_{\pm2.0}$ & $86.2_{\pm0.8}$ & $85.8_{\pm0.5}$ & $75.1_{\pm1.6}$ & $82.8_{\pm0.6}$ & $79.3_{\pm1.1}$ & $76.0_{\pm1.7}$ & $78.8_{\pm1.5}$ & $70.7_{\pm1.8}$ & $82.1_{\pm0.7}$ & $78.6_{\pm1.4}$ & $80.0_{\pm0.9}$ \\
        NLLB-\llmvec S1+S2 & $91.4_{\pm0.2}$ & $81.9_{\pm0.4}$ & $86.5_{\pm0.5}$ & $85.8_{\pm0.3}$ & $79.3_{\pm6.4}$ & $87.4_{\pm0.3}$ & $86.7_{\pm0.4}$ & $76.8_{\pm0.6}$ & $83.6_{\pm0.5}$ & $79.6_{\pm0.4}$ & $77.4_{\pm0.7}$ & $80.0_{\pm0.3}$ & $72.1_{\pm0.5}$ & $83.2_{\pm0.3}$ & $80.0_{\pm0.9}$ & $81.4_{\pm0.6}$ \\
        \hline
        \rowcolor{gray!10} \multicolumn{17}{l}{\rule{0pt}{2.5ex} \textit{Translate-Test: Translate test data to English}} \\ 
        \hline
        \llmvec NLLB-600M  & $92.5_{\pm0.3}$ & $78.2_{\pm0.7}$ & $83.3_{\pm0.5}$ & $83.8_{\pm0.8}$ & $82.8_{\pm0.8}$ & $85.8_{\pm0.8}$ & $84.1_{\pm0.9}$ & $75.7_{\pm0.6}$ & $78.9_{\pm0.5}$ & $73.2_{\pm0.9}$ & $72.7_{\pm0.7}$ & $79.2_{\pm0.7}$ & $69.5_{\pm0.8}$ & $79.7_{\pm0.5}$ & $74.9_{\pm0.5}$ & $78.7_{\pm0.7}$  \\
        \llmvec NLLB-3B    & $92.5_{\pm0.3}$ & $80.0_{\pm0.8}$ & $85.1_{\pm0.5}$ & $85.3_{\pm0.7}$ & $83.9_{\pm0.5}$ & $86.5_{\pm0.8}$ & $85.0_{\pm0.6}$ & $76.9_{\pm0.6}$ & $81.2_{\pm0.4}$ & $74.4_{\pm1.0}$ & $74.5_{\pm0.9}$ & $81.2_{\pm0.7}$ & $70.0_{\pm0.4}$ & $81.3_{\pm0.6}$ & $77.7_{\pm0.4}$ & $80.2_{\pm0.6}$   \\
        \bottomrule
    \end{tabular}
\end{adjustbox}
\caption{\textbf{XNLI (1/2).} We benchmark models on test portions of XNLI (cf. \S\ref{sec:experimental-setup}). S1 and S2 refer to self-supervised and task-specific stages of aligning NLLB with \llmvec (cf. \S\ref{sec:methodology}). FT denotes supervised fine-tuning. Reported performance is averaged over three seeds on model checkpoints that maximize performance on \textbf{source-language (\sdev) validation splits}. Subscripts denote std. deviation. Metric: accuracy.}
\label{tab:xnli-src-dev}
\end{table*}

\begin{table*}[t!]
\begin{adjustbox}{width=\textwidth,center}
    \begin{tabular}{@{\extracolsep{\fill}} l!{\vrule width \arrayrulewidth}c    !{\vrule width \arrayrulewidth}!{\vrule width \arrayrulewidth}    c!{\vrule width \arrayrulewidth}c!{\vrule width \arrayrulewidth}c!{\vrule width \arrayrulewidth}c!{\vrule width \arrayrulewidth}c!{\vrule width \arrayrulewidth}c!{\vrule width \arrayrulewidth}c!{\vrule width \arrayrulewidth}c!{\vrule width \arrayrulewidth}c!{\vrule width \arrayrulewidth}c!{\vrule width \arrayrulewidth}c!{\vrule width \arrayrulewidth}c!{\vrule width \arrayrulewidth}c!{\vrule width \arrayrulewidth}c!{\vrule width \arrayrulewidth}c}
        \toprule
         & \textbf{\textsc{EN}} & \textbf{\textsc{AR}} & \textbf{\textsc{BG}} & \textbf{\textsc{DE}} & \textbf{\textsc{EL}} & \textbf{\textsc{ES}} & \textbf{\textsc{FR}} & \textbf{\textsc{HI}} & \textbf{\textsc{RU}} & \textbf{\textsc{SW}} & \textbf{\textsc{TH}} & \textbf{\textsc{TR}} & \textbf{\textsc{UR}} & \textbf{\textsc{VI}} & \textbf{\textsc{ZH}} & \textbf{\textsc{AVG}} \\
        \hline
        \rowcolor{gray!10} \multicolumn{17}{l}{\rule{0pt}{2.5ex} \textit{Zero-Shot Cross-Lingual Transfer: Fine-tune multilingual model on English training set}}  \\ 
        \hline
        \llmvec            & $92.5_{\pm0.3}$ & $66.3_{\pm5.1}$ & $70.8_{\pm3.5}$ & $81.4_{\pm3.0}$ & $73.8_{\pm3.5}$ & $84.8_{\pm1.2}$ & $83.5_{\pm1.2}$ & $61.8_{\pm8.0}$ & $78.3_{\pm2.4}$ & $48.5_{\pm3.9}$ & $67.1_{\pm2.9}$ & $70.2_{\pm1.2}$ & $51.4_{\pm5.6}$ & $78.0_{\pm1.3}$ & $79.0_{\pm1.3}$ & $71.1_{\pm2.4}$ \\
        NLLB-Encoder       & $80.4_{\pm0.2}$ & $71.3_{\pm0.8}$ & $75.6_{\pm0.5}$ & $75.0_{\pm0.4}$ & $74.7_{\pm0.1}$ & $75.6_{\pm0.6}$ & $75.6_{\pm0.1}$ & $68.7_{\pm0.8}$ & $72.8_{\pm0.2}$ & $69.5_{\pm0.2}$ & $66.0_{\pm0.6}$ & $72.9_{\pm0.3}$ & $65.2_{\pm0.8}$ & $72.8_{\pm0.2}$ & $69.2_{\pm0.6}$ & $71.8_{\pm0.3}$ \\
        NLLB-\llmvec S1+FT & $90.7_{\pm0.4}$ & $81.2_{\pm0.6}$ & $85.5_{\pm0.9}$ & $85.0_{\pm0.7}$ & $75.0_{\pm2.2}$ & $86.5_{\pm0.5}$ & $85.8_{\pm0.5}$ & $75.8_{\pm0.9}$ & $82.8_{\pm0.6}$ & $79.4_{\pm0.9}$ & $77.0_{\pm0.6}$ & $79.1_{\pm1.4}$ & $71.1_{\pm1.6}$ & $82.4_{\pm0.4}$ & $79.5_{\pm0.4}$ & $80.4_{\pm0.4}$ \\
        NLLB-\llmvec S1+S2 & $91.4_{\pm0.2}$ & $82.2_{\pm0.8}$ & $86.2_{\pm0.4}$ & $85.9_{\pm0.4}$ & $79.8_{\pm5.8}$ & $87.5_{\pm0.7}$ & $86.8_{\pm0.6}$ & $76.9_{\pm0.7}$ & $84.0_{\pm0.3}$ & $79.9_{\pm1.0}$ & $77.8_{\pm1.0}$ & $80.0_{\pm0.4}$ & $72.2_{\pm1.1}$ & $83.4_{\pm0.4}$ & $80.8_{\pm0.4}$ & $81.7_{\pm0.5}$ \\
        \hline
        \rowcolor{gray!10} \multicolumn{17}{l}{\rule{0pt}{2.5ex} \textit{Translate-Test: Translate test data to English}} \\ 
        \hline
        \llmvec NLLB-600M  & $92.5_{\pm0.3}$ & $77.9_{\pm1.5}$ & $82.1_{\pm0.4}$ & $83.6_{\pm0.9}$ & $82.2_{\pm0.4}$ & $86.2_{\pm0.5}$ & $84.2_{\pm1.0}$ & $74.9_{\pm0.7}$ & $78.7_{\pm1.5}$ & $73.7_{\pm1.8}$ & $72.8_{\pm0.8}$ & $79.2_{\pm1.6}$ & $69.9_{\pm2.2}$ & $79.7_{\pm0.5}$ & $75.1_{\pm0.5}$  & $78.6_{\pm0.9}$  \\
        \llmvec NLLB-3B    & $92.5_{\pm0.3}$ & $79.9_{\pm1.4}$ & $84.0_{\pm0.1}$ & $85.1_{\pm1.1}$ & $83.5_{\pm0.5}$ & $87.0_{\pm0.9}$ & $85.3_{\pm1.1}$ & $76.3_{\pm0.2}$ & $80.9_{\pm1.5}$ & $75.0_{\pm2.0}$ & $74.6_{\pm0.9}$ & $81.7_{\pm1.5}$ & $70.0_{\pm1.7}$ & $81.3_{\pm0.6}$ & $77.7_{\pm0.3}$  & $80.2_{\pm0.8}$  \\
        \bottomrule
    \end{tabular}
\end{adjustbox}
\caption{\textbf{XNLI (2/2).} We benchmark models on test portions of XNLI (cf. \S\ref{sec:experimental-setup}). S1 and S2 refer to self-supervised and task-specific stages of aligning NLLB with \llmvec (cf. \S\ref{sec:methodology}). FT denotes supervised fine-tuning. Reported performance is averaged over three seeds on model checkpoints that maximize performance on \textbf{per target-language (\tdev) validation splits}. Subscripts denote std. deviation. Metric: accuracy.}
\label{tab:xnli-trg-dev}
\end{table*}

\begin{table*}[t!]
\begin{adjustbox}{width=\textwidth,center}
    \begin{tabular}{@{\extracolsep{\fill}} l!{\vrule width \arrayrulewidth}cc!{\vrule width \arrayrulewidth}cc!{\vrule width \arrayrulewidth}cc!{\vrule width \arrayrulewidth}cc!{\vrule width \arrayrulewidth}    !{\vrule width \arrayrulewidth}cc!{\vrule width \arrayrulewidth}cc!{\vrule width \arrayrulewidth}cc!{\vrule width \arrayrulewidth}cc!{\vrule width \arrayrulewidth}cc}
        \toprule
        \multicolumn{1}{c}{} & \multicolumn{2}{c}{\textbf{\textsc{AYM}}} & \multicolumn{2}{c}{\textbf{\textsc{GN}}} & \multicolumn{2}{c}{\textbf{\textsc{QUY}}} & \multicolumn{2}{c}{\textbf{\textsc{AVG}}} & \multicolumn{2}{c}{\textbf{\textsc{AZ}}} & \multicolumn{2}{c}{\textbf{\textsc{KK}}} & \multicolumn{2}{c}{\textbf{\textsc{KY}}} & \multicolumn{2}{c}{\textbf{\textsc{UZ}}} & \multicolumn{2}{c}{\textbf{\textsc{AVG}}} \\
        \cmidrule(lr){2-3} \cmidrule(lr){4-5} \cmidrule(lr){6-7} \cmidrule(lr){8-9} \cmidrule(lr){10-11} \cmidrule(lr){12-13} \cmidrule(lr){14-15} \cmidrule(lr){16-17} \cmidrule(lr){18-19}
        & \sdev & \tdev & \sdev & \tdev & \sdev & \tdev & \sdev & \tdev & \sdev & \tdev & \sdev & \tdev & \sdev & \tdev & \sdev & \tdev & \sdev & \tdev \\
        \hline
        \rowcolor{gray!10} \multicolumn{19}{l}{\rule{0pt}{2.5ex} \textit{Zero-Shot Cross-Lingual Transfer: Fine-tune multilingual model on English training set}}  \\ 
        \hline
        \llmvec         & $40.9_{\pm3.2}$ & $41.8_{\pm2.0}$ & $42.3_{\pm1.7}$ & $44.8_{\pm1.8}$ & $39.6_{\pm1.4}$ & $42.8_{\pm1.0}$ & $40.9_{\pm2.0}$ & $43.2_{\pm1.6}$ & $54.0_{\pm0.4}$ & $61.1_{\pm0.9}$ & $43.5_{\pm2.3}$ & $46.3_{\pm1.4}$ & $41.8_{\pm1.4}$ & $43.1_{\pm0.1}$ & $47.5_{\pm3.3}$ & $55.9_{\pm2.0}$ & $46.7_{\pm1.7}$ & $51.6_{\pm0.3}$ \\
        NLLB-Encoder   & $62.3_{\pm0.5}$ & $63.4_{\pm0.1}$ & $62.8_{\pm0.9}$ & $64.2_{\pm0.2}$ & $40.8_{\pm1.0}$ & $41.5_{\pm1.0}$ & $55.3_{\pm0.6}$ & $56.4_{\pm0.3}$ & $75.8_{\pm0.4}$ & $76.1_{\pm0.2}$ & $74.2_{\pm0.5}$ & $74.3_{\pm0.6}$ & $74.8_{\pm0.5}$ & $75.2_{\pm0.5}$ & $74.8_{\pm0.8}$ & $75.1_{\pm1.0}$ & $74.9_{\pm0.5}$ & $75.2_{\pm0.6}$ \\
        NLLB-\llmvec S1+FT & $60.4_{\pm2.2}$ & $62.5_{\pm1.6}$ & $68.8_{\pm2.4}$ & $69.2_{\pm1.4}$ & $59.9_{\pm1.9}$ & $61.2_{\pm1.4}$ & $63.0_{\pm1.9}$ & $64.3_{\pm1.2}$ &  $82.8_{\pm1.0}$ & $82.3_{\pm0.8}$ & $81.1_{\pm1.2}$ & $80.0_{\pm0.8}$ & $80.3_{\pm1.9}$ & $81.3_{\pm0.7}$ & $81.9_{\pm1.4}$ & $81.5_{\pm0.7}$ & $81.5_{\pm1.3}$ & $81.3_{\pm0.4}$ \\
        NLLB-\llmvec S1+S2 & $61.0_{\pm0.6}$ & $61.8_{\pm1.0}$ & $69.6_{\pm1.3}$ & $69.8_{\pm1.4}$ & $61.4_{\pm1.8}$ & $62.3_{\pm1.1}$ & $64.0_{\pm0.3}$ & $64.6_{\pm0.7}$ & $83.4_{\pm1.2}$ & $83.0_{\pm0.9}$ & $81.9_{\pm0.5}$ & $81.7_{\pm0.7}$ & $81.2_{\pm0.4}$ & $81.6_{\pm0.9}$ & $82.1_{\pm0.2}$ & $82.3_{\pm0.5}$ & $82.5_{\pm0.5}$ & $82.1_{\pm0.5}$ \\
        \hline
        \rowcolor{gray!10} \multicolumn{19}{l}{\rule{0pt}{2.5ex} \textit{Translate-Test: Translate test data to English}} \\ 
        \hline
        \llmvec NLLB-600M & $50.7_{\pm1.9}$ & $51.2_{\pm1.9}$ & $55.0_{\pm0.2}$ & $56.8_{\pm1.7}$ & $50.5_{\pm0.3}$ & $50.0_{\pm1.3}$ & $52.0_{\pm0.7}$ & $52.7_{\pm0.6}$ & $82.2_{\pm1.2}$ & $81.7_{\pm1.1}$ & $77.2_{\pm0.4}$ & $77.1_{\pm0.4}$ & $76.5_{\pm1.0}$ & $75.8_{\pm1.9}$ & $79.2_{\pm0.8}$ & $79.1_{\pm0.8}$ & $78.8_{\pm0.8}$ & $78.4_{\pm1.0}$ \\
        \llmvec NLLB-3B  & $45.2_{\pm0.4}$ & $45.0_{\pm3.3}$ & $58.0_{\pm0.9}$ & $59.3_{\pm2.1}$ & $49.6_{\pm1.0}$ & $49.3_{\pm0.5}$ & $50.9_{\pm0.4}$ & $51.2_{\pm0.7}$ & $84.9_{\pm0.8}$ & $84.9_{\pm0.9}$ & $78.5_{\pm1.1}$ & $78.9_{\pm0.9}$ & $75.8_{\pm0.5}$ & $75.1_{\pm1.2}$ & $80.4_{\pm1.1}$ & $80.8_{\pm1.1}$ & $79.9_{\pm0.9}$ & $79.9_{\pm1.0}$ \\
        \bottomrule
    \end{tabular}
\end{adjustbox}
\caption{ \textbf{AmNLI \& Kardeş-NLU.} We benchmark models on test portions of AmNLI and Kardeş-NLU (cf. \S\ref{sec:experimental-setup}). S1 and S2 refer to self-supervised and task-specific stages of aligning NLLB with \llmvec (cf. \S\ref{sec:methodology}). FT denotes supervised fine-tuning. Reported performance is averaged over three seeds on model checkpoints that maximize performance on source-language (\sdev) and per target-language (\tdev) validation splits. Subscripts denote std. deviation. Metric: accuracy. }
\label{tab:amnli-kardesnlu}
\end{table*}

\begin{table*}[t!]
\begin{adjustbox}{width=\textwidth,center}
    \begin{tabular}{@{\extracolsep{\fill}} l!{\vrule width \arrayrulewidth}c!{\vrule width \arrayrulewidth}c!{\vrule width \arrayrulewidth}c!{\vrule width \arrayrulewidth}c!{\vrule width \arrayrulewidth}c!{\vrule width \arrayrulewidth}c!{\vrule width \arrayrulewidth}c!{\vrule width \arrayrulewidth}c!{\vrule width \arrayrulewidth}c!{\vrule width \arrayrulewidth}c}
        \toprule
         & \textbf{\textsc{ENG}} & \textbf{\textsc{ACE}} & \textbf{\textsc{BAN}} & \textbf{\textsc{BJN}} & \textbf{\textsc{BUG}} & \textbf{\textsc{IND}} & \textbf{\textsc{JAV}} & \textbf{\textsc{MIN}} & \textbf{\textsc{SUN}} & \textbf{\textsc{AVG}} \\
        \hline
        \rowcolor{gray!10} \multicolumn{11}{l}{\rule{0pt}{2.5ex} \textit{Zero-Shot Cross-Lingual Transfer: Fine-tune multilingual model on English training set}}  \\ 
        \hline
        \llmvec            & $91.3_{\pm0.5}$ & $41.6_{\pm14.5}$ & $45.1_{\pm17.6}$ & $56.9_{\pm13.9}$ & $30.5_{\pm16.7}$ & $83.3_{\pm2.4}$ & $54.4_{\pm14.8}$ & $56.5_{\pm16.8}$ & $43.6_{\pm17.9}$ & $51.5_{\pm14.0}$ \\
        NLLB-Encoder       & $86.9_{\pm2.1}$ & $80.3_{\pm0.8}$ & $76.9_{\pm2.0}$ & $83.8_{\pm1.0}$ & $67.4_{\pm2.7}$ & $86.4_{\pm0.7}$ & $83.6_{\pm0.8}$ & $80.1_{\pm0.5}$ & $80.7_{\pm0.4}$ & $79.9_{\pm0.3}$ \\ 
        NLLB-\llmvec S1+FT & $90.8_{\pm0.6}$ & $73.9_{\pm4.3}$ & $70.6_{\pm2.4}$ & $79.1_{\pm2.5}$ & $53.9_{\pm9.3}$ & $86.7_{\pm2.3}$ & $81.0_{\pm2.4}$ & $72.6_{\pm6.3}$ & $78.4_{\pm2.7}$ & $74.5_{\pm3.2}$ \\
        NLLB-\llmvec S1+S2 & $92.2_{\pm0.5}$ & $81.5_{\pm2.5}$ & $74.8_{\pm4.1}$ & $82.3_{\pm2.6}$ & $67.1_{\pm1.7}$ & $89.3_{\pm0.6}$ & $86.4_{\pm1.8}$ & $80.6_{\pm3.3}$ & $83.1_{\pm3.1}$ & $80.6_{\pm2.3}$ \\
        \hline
        \rowcolor{gray!10} \multicolumn{11}{l}{\rule{0pt}{2.5ex} \textit{Translate-Test: Translate test data to English}} \\ 
        \hline
        \llmvec NLLB-600M  & $91.3_{\pm0.5}$ & $74.2_{\pm2.2}$ & $72.1_{\pm1.6}$ & $79.1_{\pm1.8}$ & $71.3_{\pm3.7}$ & $86.7_{\pm1.2}$ & $79.4_{\pm1.9}$ & $78.7_{\pm1.6}$ & $81.8_{\pm1.2}$ & $77.9_{\pm1.4}$ \\
        \llmvec NLLB-3B    & $91.3_{\pm0.5}$ & $77.7_{\pm1.4}$ & $75.6_{\pm1.3}$ & $83.8_{\pm1.5}$ & $71.5_{\pm4.3}$ & $88.7_{\pm0.6}$ & $84.3_{\pm1.3}$ & $82.1_{\pm0.4}$ & $86.1_{\pm1.0}$ & $81.2_{\pm0.8}$ \\
        \bottomrule
    \end{tabular}
\end{adjustbox}
\caption{\textbf{NusaX.} We benchmark models on test portions of NusaX (cf. \S\ref{sec:experimental-setup}). S1 and S2 refer to self-supervised and task-specific stages of aligning NLLB with \llmvec (cf. \S\ref{sec:methodology}). FT denotes supervised fine-tuning. Reported performance is averaged over three seeds on model checkpoints that maximize performance on \textbf{per target-language (\tdev) validation splits}. Subscripts denote std. deviation. Metric: macro-F1.}
\label{tab:nusax}
\end{table*}

\begin{table*}[t!]
\begin{adjustbox}{width=\textwidth,center}
    \begin{tabular}{@{\extracolsep{\fill}} l!{\vrule width \arrayrulewidth}cccccc}
        \toprule
        \multicolumn{1}{c}{} & \multicolumn{4}{c}{\textbf{\textsc{Zero-Shot Cross-Lingual Transfer}}} & \multicolumn{2}{c}{\textbf{\textsc{Translate-Test}}} \\
        \cmidrule(lr){2-5} \cmidrule(lr){6-7}
         & \llmvec & NLLB-Encoder & NLLB-\llmvec S1+FT  & NLLB-\llmvec S1+S2 & \llmvec NLLB-600M & \llmvec NLLB-3B \\
        \hline
        \texttt{eng\_Latn} & $92.5_{\pm0.3}$ & $80.4_{\pm0.2}$ & $90.0_{\pm0.7}$ & $91.4_{\pm0.2}$ & $92.5_{\pm0.3}$ & $92.5_{\pm0.3}$\\ \hline\hline
        AVG       & $48.2_{\pm3.2}$  & $30.4_{\pm0.4}$ & $60.2_{\pm0.5}$ & $62.6_{\pm0.5}$ & $60.7_{\pm0.8}$ & $64.2_{\pm0.7}$\\ \hline
        
        \texttt{acm\_Arab} & $52.3_{\pm4.1}$  & $30.1_{\pm0.2}$ & $55.8_{\pm1.2}$ & $56.7_{\pm0.6}$ & $62.9_{\pm0.6}$ & $60.2_{\pm0.3}$ \\
        \texttt{afr\_Latn} & $66.9_{\pm8.5}$  & $33.3_{\pm0.2}$ & $76.3_{\pm1.4}$ & $80.1_{\pm1.1}$ & $79.4_{\pm0.5}$ & $79.8_{\pm0.2}$ \\
        \texttt{als\_Latn} & $47.2_{\pm7.5}$  & $28.9_{\pm0.4}$ & $70.8_{\pm2.5}$ & $72.9_{\pm1.0}$ & $71.1_{\pm1.0}$ & $75.7_{\pm1.3}$ \\
        \texttt{amh\_Ethi} & $27.7_{\pm0.9}$  & $30.8_{\pm1.1}$ & $50.0_{\pm0.6}$ & $50.4_{\pm1.5}$ & $50.9_{\pm1.4}$ & $59.0_{\pm1.0}$ \\
        \texttt{apc\_Arab} & $52.1_{\pm2.6}$  & $30.6_{\pm1.6}$ & $58.8_{\pm0.9}$ & $60.7_{\pm0.6}$ & $66.3_{\pm0.8}$ & $65.1_{\pm0.6}$ \\
        \texttt{arb\_Arab} & $68.5_{\pm6.3}$  & $28.5_{\pm1.1}$ & $65.7_{\pm0.9}$ & $68.1_{\pm1.5}$ & $75.2_{\pm1.0}$ & $75.9_{\pm0.6}$ \\
        \texttt{ars\_Arab} & $55.2_{\pm4.8}$  & $29.9_{\pm1.6}$ & $58.4_{\pm0.5}$ & $59.6_{\pm1.1}$ & $66.9_{\pm1.3}$ & $61.3_{\pm0.2}$ \\
        \texttt{ary\_Arab} & $45.2_{\pm3.2}$  & $31.0_{\pm0.5}$ & $46.5_{\pm1.1}$ & $48.3_{\pm0.8}$ & $51.4_{\pm2.3}$ & $54.9_{\pm0.8}$ \\
        \texttt{arz\_Arab} & $50.1_{\pm6.5}$  & $31.2_{\pm1.6}$ & $56.8_{\pm0.7}$ & $59.7_{\pm0.8}$ & $68.6_{\pm1.4}$ & $67.4_{\pm0.4}$ \\
        \texttt{asm\_Beng} & $29.1_{\pm1.8}$  & $29.1_{\pm0.5}$ & $50.9_{\pm2.5}$ & $54.0_{\pm1.1}$ & $52.5_{\pm1.3}$ & $60.3_{\pm0.2}$ \\
        \texttt{azj\_Latn} & $44.5_{\pm7.0}$  & $29.7_{\pm0.6}$ & $53.7_{\pm0.3}$ & $56.7_{\pm0.8}$ & $64.2_{\pm0.8}$ & $66.7_{\pm1.0}$ \\
        \texttt{bam\_Latn} & $32.1_{\pm1.4}$  & $29.3_{\pm0.2}$ & $41.1_{\pm0.9}$ & $39.5_{\pm1.2}$ & $36.0_{\pm0.8}$ & $37.3_{\pm1.3}$ \\
        \texttt{ben\_Beng} & $31.9_{\pm0.8}$  & $28.5_{\pm0.8}$ & $58.1_{\pm2.4}$ & $62.5_{\pm0.8}$ & $63.5_{\pm0.8}$ & $65.9_{\pm0.7}$ \\
        \texttt{ben\_Latn} & $34.9_{\pm0.7}$  & $27.7_{\pm0.1}$ & $27.0_{\pm0.9}$ & $28.2_{\pm1.2}$ & $29.6_{\pm0.7}$ & $25.5_{\pm1.5}$ \\
        \texttt{bod\_Tibt} & $26.7_{\pm1.9}$  & $28.0_{\pm1.5}$ & $33.5_{\pm1.3}$ & $33.9_{\pm1.1}$ & $29.7_{\pm1.7}$ & $35.9_{\pm1.9}$ \\
        \texttt{bul\_Cyrl} & $73.4_{\pm7.5}$  & $30.8_{\pm1.2}$ & $75.3_{\pm1.2}$ & $77.9_{\pm0.1}$ & $70.0_{\pm1.7}$ & $77.9_{\pm0.5}$ \\
        \texttt{cat\_Latn} & $77.6_{\pm6.0}$  & $33.7_{\pm1.0}$ & $79.1_{\pm1.4}$ & $82.3_{\pm0.9}$ & $74.3_{\pm1.5}$ & $79.5_{\pm0.8}$ \\
        \texttt{ceb\_Latn} & $43.8_{\pm3.7}$  & $29.1_{\pm1.2}$ & $67.0_{\pm1.6}$ & $70.7_{\pm0.7}$ & $66.9_{\pm1.9}$ & $73.2_{\pm1.5}$ \\
        \texttt{ces\_Latn} & $73.7_{\pm7.6}$  & $30.2_{\pm1.4}$ & $71.3_{\pm0.6}$ & $75.9_{\pm1.5}$ & $70.8_{\pm0.9}$ & $76.8_{\pm1.0}$ \\
        \texttt{ckb\_Arab} & $33.2_{\pm0.9}$  & $28.7_{\pm0.6}$ & $58.3_{\pm0.5}$ & $59.8_{\pm1.2}$ & $62.7_{\pm1.3}$ & $65.1_{\pm1.0}$ \\
        \texttt{dan\_Latn} & $73.0_{\pm7.3}$  & $32.2_{\pm0.7}$ & $81.0_{\pm0.8}$ & $83.5_{\pm0.7}$ & $73.8_{\pm1.7}$ & $79.2_{\pm1.3}$ \\
        \texttt{deu\_Latn} & $85.1_{\pm2.6}$  & $33.8_{\pm0.7}$ & $76.0_{\pm0.7}$ & $78.1_{\pm0.9}$ & $76.1_{\pm1.0}$ & $80.1_{\pm0.2}$ \\
        \texttt{ell\_Grek} & $74.8_{\pm6.7}$  & $28.7_{\pm0.9}$ & $62.2_{\pm1.6}$ & $67.5_{\pm0.2}$ & $70.7_{\pm0.3}$ & $76.5_{\pm0.8}$ \\
        \texttt{est\_Latn} & $46.1_{\pm8.0}$  & $30.6_{\pm0.3}$ & $66.0_{\pm2.0}$ & $70.1_{\pm1.0}$ & $64.2_{\pm1.4}$ & $71.4_{\pm1.1}$ \\
        \texttt{eus\_Latn} & $45.4_{\pm6.4}$  & $31.0_{\pm1.2}$ & $63.1_{\pm0.6}$ & $66.7_{\pm1.1}$ & $72.6_{\pm0.6}$ & $75.9_{\pm0.7}$ \\
        \texttt{fin\_Latn} & $55.1_{\pm9.7}$  & $31.1_{\pm1.1}$ & $69.0_{\pm0.7}$ & $73.0_{\pm0.7}$ & $67.3_{\pm1.5}$ & $77.7_{\pm0.4}$ \\
        \texttt{fra\_Latn} & $88.0_{\pm1.5}$  & $31.6_{\pm0.5}$ & $79.1_{\pm0.5}$ & $82.4_{\pm1.0}$ & $80.0_{\pm1.3}$ & $82.6_{\pm0.6}$ \\
        \texttt{fuv\_Latn} & $28.5_{\pm0.4}$  & $28.0_{\pm1.2}$ & $29.5_{\pm0.8}$ & $28.1_{\pm0.6}$ & $27.9_{\pm0.5}$ & $26.9_{\pm1.5}$ \\
        \texttt{gaz\_Latn} & $31.4_{\pm0.6}$  & $29.4_{\pm0.3}$ & $41.6_{\pm1.7}$ & $42.8_{\pm0.7}$ & $45.0_{\pm0.3}$ & $48.6_{\pm0.6}$ \\
        \texttt{grn\_Latn} & $37.2_{\pm0.9}$  & $31.0_{\pm1.2}$ & $52.1_{\pm0.4}$ & $52.5_{\pm0.9}$ & $47.3_{\pm1.4}$ & $54.1_{\pm0.8}$ \\
        \texttt{guj\_Gujr} & $27.9_{\pm0.4}$  & $30.5_{\pm1.3}$ & $52.9_{\pm0.7}$ & $55.9_{\pm1.7}$ & $62.6_{\pm1.0}$ & $64.7_{\pm0.6}$ \\
        \texttt{hat\_Latn} & $38.2_{\pm2.5}$  & $29.1_{\pm1.1}$ & $63.3_{\pm1.3}$ & $67.4_{\pm0.9}$ & $65.9_{\pm0.9}$ & $71.5_{\pm0.7}$ \\
        \texttt{hau\_Latn} & $32.0_{\pm0.8}$  & $28.4_{\pm0.6}$ & $58.3_{\pm1.5}$ & $62.1_{\pm0.4}$ & $59.4_{\pm2.2}$ & $59.9_{\pm1.1}$ \\
        \texttt{heb\_Hebr} & $39.8_{\pm6.8}$  & $32.8_{\pm0.4}$ & $64.0_{\pm1.7}$ & $66.6_{\pm0.3}$ & $68.8_{\pm1.6}$ & $71.7_{\pm0.7}$ \\
        \texttt{hin\_Deva} & $55.1_{\pm5.7}$  & $28.6_{\pm1.0}$ & $62.4_{\pm1.6}$ & $65.7_{\pm1.0}$ & $70.6_{\pm0.6}$ & $73.0_{\pm1.5}$ \\
        \texttt{hrv\_Latn} & $63.4_{\pm9.5}$  & $31.5_{\pm1.1}$ & $73.1_{\pm1.0}$ & $77.4_{\pm0.5}$ & $69.7_{\pm1.2}$ & $73.3_{\pm0.5}$ \\
        \texttt{hun\_Latn} & $62.4_{\pm2.6}$  & $30.9_{\pm1.0}$ & $67.8_{\pm0.7}$ & $71.6_{\pm0.5}$ & $66.9_{\pm0.5}$ & $72.7_{\pm1.5}$ \\
        \texttt{hye\_Armn} & $27.8_{\pm0.6}$  & $28.4_{\pm0.8}$ & $56.6_{\pm1.5}$ & $58.6_{\pm0.6}$ & $52.0_{\pm1.3}$ & $61.5_{\pm1.7}$ \\
        \texttt{ibo\_Latn} & $31.0_{\pm0.9}$  & $30.2_{\pm1.2}$ & $48.8_{\pm1.1}$ & $49.2_{\pm0.3}$ & $47.0_{\pm1.6}$ & $51.7_{\pm1.8}$ \\
        \texttt{ilo\_Latn} & $38.6_{\pm1.2}$  & $29.6_{\pm1.9}$ & $62.6_{\pm0.9}$ & $66.7_{\pm1.3}$ & $61.9_{\pm0.8}$ & $67.7_{\pm0.7}$ \\
        \texttt{ind\_Latn} & $73.3_{\pm6.7}$  & $30.6_{\pm0.6}$ & $79.4_{\pm1.3}$ & $82.5_{\pm0.6}$ & $74.7_{\pm0.7}$ & $76.5_{\pm0.4}$ \\
        \texttt{isl\_Latn} & $44.2_{\pm5.4}$  & $28.3_{\pm1.5}$ & $61.6_{\pm0.3}$ & $65.0_{\pm1.2}$ & $56.3_{\pm0.8}$ & $57.9_{\pm1.1}$ \\
        \texttt{ita\_Latn} & $85.9_{\pm1.5}$  & $31.9_{\pm1.7}$ & $79.8_{\pm0.0}$ & $82.4_{\pm0.5}$ & $74.0_{\pm1.6}$ & $78.2_{\pm1.4}$ \\
        \texttt{jav\_Latn} & $45.2_{\pm5.3}$  & $29.4_{\pm0.9}$ & $71.4_{\pm1.0}$ & $74.2_{\pm0.5}$ & $56.6_{\pm1.0}$ & $58.1_{\pm0.8}$ \\
        \texttt{jpn\_Jpan} & $77.3_{\pm2.4}$  & $30.4_{\pm1.1}$ & $65.5_{\pm2.2}$ & $67.1_{\pm0.9}$ & $60.7_{\pm0.8}$ & $65.1_{\pm0.7}$ \\
        \texttt{kac\_Latn} & $32.7_{\pm0.5}$  & $29.6_{\pm0.6}$ & $39.1_{\pm0.4}$ & $40.9_{\pm1.2}$ & $37.0_{\pm0.3}$ & $39.7_{\pm1.3}$ \\
        \texttt{kan\_Knda} & $28.6_{\pm0.4}$  & $30.3_{\pm1.0}$ & $55.7_{\pm1.4}$ & $56.9_{\pm1.3}$ & $62.4_{\pm1.3}$ & $65.6_{\pm1.0}$ \\
        \texttt{kat\_Geor} & $27.1_{\pm1.6}$  & $27.4_{\pm0.4}$ & $50.4_{\pm1.9}$ & $51.1_{\pm0.5}$ & $50.3_{\pm2.0}$ & $56.7_{\pm2.0}$ \\
        \texttt{kaz\_Cyrl} & $40.7_{\pm4.7}$  & $29.6_{\pm0.6}$ & $55.9_{\pm1.3}$ & $59.3_{\pm0.7}$ & $65.3_{\pm0.5}$ & $69.1_{\pm0.4}$ \\
        \texttt{kea\_Latn} & $43.8_{\pm1.3}$  & $31.0_{\pm0.3}$ & $61.2_{\pm1.3}$ & $65.6_{\pm1.5}$ & $59.7_{\pm0.8}$ & $62.8_{\pm0.5}$ \\
        \texttt{khk\_Cyrl} & $33.9_{\pm2.0}$  & $28.4_{\pm0.5}$ & $44.4_{\pm1.5}$ & $44.8_{\pm1.2}$ & $48.1_{\pm0.5}$ & $52.5_{\pm1.7}$ \\
        \texttt{khm\_Khmr} & $29.6_{\pm2.2}$  & $29.8_{\pm0.2}$ & $47.3_{\pm0.8}$ & $51.6_{\pm0.8}$ & $44.4_{\pm0.8}$ & $47.7_{\pm1.6}$ \\
        \texttt{kin\_Latn} & $36.2_{\pm1.0}$  & $29.3_{\pm0.2}$ & $55.3_{\pm0.7}$ & $57.1_{\pm0.8}$ & $55.0_{\pm1.0}$ & $58.3_{\pm0.8}$ \\
        \texttt{kir\_Cyrl} & $40.7_{\pm3.0}$  & $30.6_{\pm0.9}$ & $56.6_{\pm1.5}$ & $58.7_{\pm0.8}$ & $62.9_{\pm1.3}$ & $66.5_{\pm1.7}$ \\
        \texttt{kor\_Hang} & $77.5_{\pm3.1}$  & $32.3_{\pm0.8}$ & $61.6_{\pm2.1}$ & $62.3_{\pm0.7}$ & $67.9_{\pm2.0}$ & $69.0_{\pm0.7}$ \\
        \texttt{lao\_Laoo} & $28.4_{\pm2.2}$  & $30.2_{\pm0.8}$ & $54.8_{\pm2.0}$ & $58.4_{\pm1.5}$ & $51.0_{\pm0.8}$ & $51.6_{\pm1.0}$ \\
        \texttt{lin\_Latn} & $33.2_{\pm1.3}$  & $28.4_{\pm0.9}$ & $53.8_{\pm1.6}$ & $57.0_{\pm0.8}$ & $52.1_{\pm0.6}$ & $57.3_{\pm1.1}$ \\
        \texttt{lit\_Latn} & $49.3_{\pm5.3}$  & $31.2_{\pm1.6}$ & $68.7_{\pm1.3}$ & $72.3_{\pm0.5}$ & $62.1_{\pm0.8}$ & $68.1_{\pm0.9}$ \\
        \texttt{lug\_Latn} & $31.1_{\pm1.6}$  & $28.3_{\pm0.6}$ & $44.6_{\pm1.3}$ & $47.2_{\pm0.3}$ & $42.6_{\pm1.0}$ & $45.6_{\pm0.9}$ \\
        \texttt{luo\_Latn} & $31.6_{\pm1.8}$  & $28.6_{\pm0.6}$ & $45.2_{\pm0.6}$ & $45.4_{\pm0.9}$ & $37.6_{\pm0.8}$ & $42.6_{\pm0.2}$ \\
        \texttt{lvs\_Latn} & $46.7_{\pm3.9}$  & $29.6_{\pm1.6}$ & $68.3_{\pm0.4}$ & $70.6_{\pm1.3}$ & $59.0_{\pm1.2}$ & $68.6_{\pm2.1}$ \\
         
        \bottomrule
    \end{tabular}
\end{adjustbox}
\caption{\textbf{Belebele (1/2).} We benchmark models on test portions of Belebele (cf. \S\ref{sec:experimental-setup}). S1 and S2 refer to self-supervised and task-specific stages of aligning NLLB with \llmvec (cf. \S\ref{sec:methodology}). FT denotes supervised fine-tuning. Reported performance is averaged over three seeds on model checkpoints that maximize performance on source-language (\sdev) validation splits. Subscripts denote std. deviation. Metric: accuracy.}
\label{tab:belebele-full-1}
\end{table*}

\begin{table*}[t!]
\begin{adjustbox}{width=\textwidth,center}
    \begin{tabular}{@{\extracolsep{\fill}} l!{\vrule width \arrayrulewidth}cccccc}
        \toprule
        \multicolumn{1}{c}{} & \multicolumn{4}{c}{\textbf{\textsc{Zero-Shot Cross-Lingual Transfer}}} & \multicolumn{2}{c}{\textbf{\textsc{Translate-Test}}} \\
        \cmidrule(lr){2-5} \cmidrule(lr){6-7}
         & \llmvec & NLLB-Encoder & NLLB-\llmvec S1+FT  & NLLB-\llmvec S1+S2 & \llmvec NLLB-600M & \llmvec NLLB-3B \\
        \hline
        \texttt{eng\_Latn} & $92.5_{\pm0.3}$ & $80.4_{\pm0.2}$ & $90.0_{\pm0.7}$ & $91.4_{\pm0.2}$ & $92.5_{\pm0.3}$ & $92.5_{\pm0.3}$\\ \hline\hline
        AVG       & $48.2_{\pm3.2}$  & $30.4_{\pm0.4}$ & $60.2_{\pm0.5}$ & $62.6_{\pm0.5}$ & $60.7_{\pm0.8}$ & $64.2_{\pm0.7}$\\ \hline
        
        \texttt{mal\_Mlym} & $28.7_{\pm0.4}$  & $30.0_{\pm0.9}$ & $49.5_{\pm2.4}$ & $49.9_{\pm0.4}$ & $66.9_{\pm1.2}$ & $65.1_{\pm0.8}$ \\
        \texttt{mar\_Deva} & $42.8_{\pm5.7}$  & $32.4_{\pm1.5}$ & $59.1_{\pm1.7}$ & $60.8_{\pm0.5}$ & $64.3_{\pm0.8}$ & $63.7_{\pm0.7}$ \\
        \texttt{mkd\_Cyrl} & $64.4_{\pm6.7}$  & $29.3_{\pm1.1}$ & $72.6_{\pm0.3}$ & $74.3_{\pm0.6}$ & $68.4_{\pm0.9}$ & $72.6_{\pm0.6}$ \\
        \texttt{mlt\_Latn} & $41.1_{\pm5.8}$  & $30.0_{\pm0.6}$ & $62.8_{\pm0.6}$ & $67.7_{\pm1.7}$ & $67.9_{\pm0.4}$ & $67.9_{\pm1.2}$ \\
        \texttt{mri\_Latn} & $31.7_{\pm1.6}$  & $26.8_{\pm1.0}$ & $46.2_{\pm0.7}$ & $47.2_{\pm1.4}$ & $49.3_{\pm1.2}$ & $52.1_{\pm0.3}$ \\
        \texttt{mya\_Mymr} & $28.1_{\pm0.9}$  & $28.9_{\pm0.4}$ & $44.8_{\pm0.8}$ & $47.4_{\pm0.5}$ & $41.8_{\pm0.8}$ & $46.1_{\pm0.8}$ \\
        \texttt{nld\_Latn} & $79.1_{\pm5.2}$  & $31.1_{\pm0.8}$ & $78.1_{\pm1.4}$ & $81.0_{\pm0.6}$ & $74.8_{\pm0.3}$ & $78.6_{\pm0.9}$ \\
        \texttt{nob\_Latn} & $73.7_{\pm7.5}$  & $32.5_{\pm0.3}$ & $81.7_{\pm1.0}$ & $84.6_{\pm0.2}$ & $75.8_{\pm0.5}$ & $79.5_{\pm1.1}$ \\
        \texttt{npi\_Deva} & $42.1_{\pm2.8}$  & $28.1_{\pm0.6}$ & $59.4_{\pm1.7}$ & $60.8_{\pm1.6}$ & $55.8_{\pm0.4}$ & $56.8_{\pm1.8}$ \\
        \texttt{nso\_Latn} & $32.3_{\pm0.7}$  & $29.3_{\pm2.2}$ & $57.6_{\pm0.1}$ & $60.0_{\pm1.5}$ & $60.0_{\pm1.4}$ & $63.6_{\pm1.6}$ \\
        \texttt{nya\_Latn} & $30.6_{\pm1.2}$  & $27.0_{\pm0.6}$ & $52.2_{\pm1.0}$ & $54.5_{\pm1.4}$ & $48.4_{\pm1.4}$ & $53.2_{\pm0.4}$ \\
        \texttt{ory\_Orya} & $27.7_{\pm1.7}$  & $30.4_{\pm0.9}$ & $56.0_{\pm1.2}$ & $57.1_{\pm1.2}$ & $66.5_{\pm1.3}$ & $71.3_{\pm0.8}$ \\
        \texttt{pan\_Guru} & $28.6_{\pm0.7}$  & $29.9_{\pm0.6}$ & $55.4_{\pm2.5}$ & $56.5_{\pm2.1}$ & $64.8_{\pm1.1}$ & $66.0_{\pm1.1}$ \\
        \texttt{pbt\_Arab} & $35.7_{\pm4.3}$  & $30.3_{\pm0.8}$ & $48.9_{\pm1.1}$ & $49.9_{\pm1.4}$ & $60.3_{\pm0.6}$ & $61.9_{\pm1.5}$ \\
        \texttt{pes\_Arab} & $71.3_{\pm4.7}$  & $31.6_{\pm1.1}$ & $69.9_{\pm1.0}$ & $71.4_{\pm0.3}$ & $67.4_{\pm0.4}$ & $70.7_{\pm0.8}$ \\
        \texttt{plt\_Latn} & $34.4_{\pm1.0}$  & $29.0_{\pm0.5}$ & $61.9_{\pm1.2}$ & $64.9_{\pm0.7}$ & $63.9_{\pm0.8}$ & $66.9_{\pm1.5}$ \\
        \texttt{pol\_Latn} & $69.4_{\pm7.0}$  & $30.8_{\pm1.0}$ & $67.6_{\pm0.9}$ & $71.0_{\pm0.8}$ & $69.3_{\pm1.5}$ & $75.4_{\pm1.4}$ \\
        \texttt{por\_Latn} & $87.0_{\pm1.5}$  & $32.9_{\pm1.0}$ & $82.1_{\pm1.7}$ & $84.1_{\pm0.7}$ & $77.7_{\pm1.1}$ & $79.7_{\pm1.5}$ \\
        \texttt{ron\_Latn} & $74.7_{\pm7.5}$  & $31.4_{\pm1.2}$ & $76.2_{\pm1.1}$ & $79.1_{\pm1.0}$ & $72.1_{\pm1.7}$ & $76.1_{\pm0.7}$ \\
        \texttt{rus\_Cyrl} & $85.9_{\pm1.9}$  & $32.6_{\pm0.3}$ & $75.4_{\pm0.3}$ & $79.1_{\pm1.2}$ & $71.8_{\pm1.2}$ & $80.7_{\pm0.6}$ \\
        \texttt{shn\_Mymr} & $26.7_{\pm0.7}$  & $25.9_{\pm1.2}$ & $34.9_{\pm2.7}$ & $37.3_{\pm0.5}$ & $34.8_{\pm1.4}$ & $36.1_{\pm0.6}$ \\
        \texttt{sin\_Sinh} & $29.6_{\pm0.3}$  & $28.1_{\pm1.1}$ & $42.0_{\pm1.8}$ & $43.8_{\pm1.6}$ & $55.9_{\pm0.5}$ & $58.7_{\pm1.4}$ \\
        \texttt{slk\_Latn} & $62.5_{\pm7.8}$  & $31.7_{\pm0.5}$ & $72.4_{\pm1.1}$ & $76.0_{\pm0.8}$ & $69.2_{\pm1.2}$ & $75.1_{\pm0.6}$ \\
        \texttt{slv\_Latn} & $54.4_{\pm7.4}$  & $31.4_{\pm0.6}$ & $72.7_{\pm1.3}$ & $76.0_{\pm0.6}$ & $68.3_{\pm1.4}$ & $75.0_{\pm1.7}$ \\
        \texttt{sna\_Latn} & $34.4_{\pm1.6}$  & $29.5_{\pm1.1}$ & $59.6_{\pm1.1}$ & $61.1_{\pm0.9}$ & $55.0_{\pm0.3}$ & $61.3_{\pm0.4}$ \\
        \texttt{snd\_Arab} & $36.4_{\pm1.1}$  & $30.1_{\pm1.7}$ & $54.5_{\pm0.7}$ & $56.6_{\pm0.3}$ & $63.8_{\pm0.6}$ & $60.3_{\pm1.5}$ \\
        \texttt{som\_Latn} & $31.2_{\pm0.8}$  & $26.8_{\pm0.9}$ & $48.4_{\pm0.9}$ & $51.2_{\pm1.1}$ & $54.3_{\pm0.7}$ & $55.4_{\pm0.9}$ \\
        \texttt{sot\_Latn} & $31.4_{\pm1.6}$  & $27.4_{\pm1.4}$ & $54.8_{\pm1.3}$ & $57.9_{\pm1.4}$ & $61.0_{\pm0.4}$ & $64.4_{\pm1.2}$ \\
        \texttt{spa\_Latn} & $87.4_{\pm1.7}$  & $30.8_{\pm0.9}$ & $82.3_{\pm1.8}$ & $85.6_{\pm0.4}$ & $74.0_{\pm0.7}$ & $79.5_{\pm1.1}$ \\
        \texttt{srp\_Cyrl} & $65.4_{\pm0.3}$  & $30.9_{\pm1.0}$ & $72.9_{\pm1.2}$ & $75.5_{\pm1.0}$ & $66.7_{\pm1.4}$ & $69.4_{\pm1.4}$ \\
        \texttt{ssw\_Latn} & $31.5_{\pm1.4}$  & $29.3_{\pm1.0}$ & $50.2_{\pm0.8}$ & $52.4_{\pm2.1}$ & $50.9_{\pm1.4}$ & $57.9_{\pm1.2}$ \\
        \texttt{sun\_Latn} & $39.6_{\pm2.7}$  & $30.2_{\pm1.5}$ & $66.3_{\pm0.8}$ & $70.1_{\pm0.6}$ & $58.7_{\pm1.4}$ & $61.1_{\pm1.3}$ \\
        \texttt{swe\_Latn} & $76.9_{\pm6.2}$  & $31.4_{\pm1.3}$ & $79.7_{\pm1.1}$ & $82.8_{\pm0.4}$ & $71.7_{\pm1.3}$ & $78.1_{\pm0.3}$ \\
        \texttt{swh\_Latn} & $42.5_{\pm2.5}$  & $30.4_{\pm1.1}$ & $73.4_{\pm0.7}$ & $75.5_{\pm0.9}$ & $69.0_{\pm1.6}$ & $75.7_{\pm1.1}$ \\
        \texttt{tam\_Taml} & $28.6_{\pm0.7}$  & $32.2_{\pm0.5}$ & $51.7_{\pm2.0}$ & $54.8_{\pm0.9}$ & $62.0_{\pm1.4}$ & $60.5_{\pm0.2}$ \\
        \texttt{tel\_Telu} & $28.2_{\pm1.3}$  & $30.6_{\pm1.1}$ & $52.0_{\pm0.9}$ & $54.6_{\pm1.0}$ & $59.6_{\pm1.0}$ & $61.6_{\pm1.0}$ \\
        \texttt{tgk\_Cyrl} & $40.6_{\pm3.6}$  & $29.2_{\pm1.0}$ & $53.5_{\pm0.7}$ & $54.7_{\pm0.6}$ & $58.0_{\pm1.5}$ & $63.9_{\pm1.1}$ \\
        \texttt{tgl\_Latn} & $52.7_{\pm8.5}$  & $29.1_{\pm0.4}$ & $68.9_{\pm1.3}$ & $72.2_{\pm1.3}$ & $71.4_{\pm1.7}$ & $71.1_{\pm1.4}$ \\
        \texttt{tha\_Thai} & $69.4_{\pm5.0}$  & $31.5_{\pm1.1}$ & $56.3_{\pm1.3}$ & $58.8_{\pm0.6}$ & $58.2_{\pm1.5}$ & $61.7_{\pm0.4}$ \\
        \texttt{tir\_Ethi} & $27.1_{\pm1.7}$  & $28.7_{\pm0.7}$ & $42.4_{\pm1.0}$ & $44.6_{\pm1.8}$ & $42.1_{\pm2.7}$ & $47.7_{\pm0.9}$ \\
        \texttt{tsn\_Latn} & $33.0_{\pm1.2}$  & $27.7_{\pm1.7}$ & $54.3_{\pm0.3}$ & $56.6_{\pm0.6}$ & $54.6_{\pm1.8}$ & $61.1_{\pm1.7}$ \\
        \texttt{tso\_Latn} & $36.0_{\pm0.5}$  & $29.1_{\pm0.3}$ & $62.6_{\pm0.7}$ & $64.5_{\pm1.7}$ & $55.8_{\pm2.6}$ & $64.3_{\pm0.8}$ \\
        \texttt{tur\_Latn} & $66.7_{\pm8.6}$  & $30.9_{\pm1.0}$ & $61.3_{\pm1.0}$ & $64.0_{\pm1.1}$ & $68.6_{\pm1.5}$ & $75.7_{\pm1.2}$ \\
        \texttt{ukr\_Cyrl} & $77.9_{\pm3.1}$  & $30.9_{\pm0.9}$ & $71.5_{\pm0.6}$ & $74.1_{\pm0.8}$ & $70.6_{\pm0.8}$ & $73.4_{\pm0.8}$ \\
        \texttt{urd\_Arab} & $44.4_{\pm6.3}$  & $31.4_{\pm0.8}$ & $62.1_{\pm1.2}$ & $63.6_{\pm1.6}$ & $66.0_{\pm0.5}$ & $67.4_{\pm1.1}$ \\
        \texttt{uzn\_Latn} & $44.9_{\pm5.6}$  & $30.0_{\pm0.9}$ & $58.6_{\pm0.7}$ & $61.0_{\pm0.8}$ & $70.0_{\pm1.1}$ & $75.0_{\pm1.7}$ \\
        \texttt{vie\_Latn} & $82.1_{\pm1.1}$  & $28.2_{\pm0.5}$ & $72.4_{\pm1.9}$ & $76.4_{\pm0.3}$ & $71.7_{\pm0.1}$ & $73.4_{\pm0.6}$ \\
        \texttt{war\_Latn} & $48.2_{\pm3.8}$  & $29.8_{\pm0.4}$ & $66.6_{\pm0.4}$ & $68.6_{\pm1.1}$ & $69.4_{\pm1.2}$ & $73.0_{\pm0.2}$ \\
        \texttt{wol\_Latn} & $31.1_{\pm0.6}$  & $27.7_{\pm0.6}$ & $36.1_{\pm0.9}$ & $36.9_{\pm0.7}$ & $35.2_{\pm1.2}$ & $33.7_{\pm1.3}$ \\
        \texttt{xho\_Latn} & $32.4_{\pm0.7}$  & $28.3_{\pm0.5}$ & $59.6_{\pm0.5}$ & $61.2_{\pm0.8}$ & $59.1_{\pm0.9}$ & $64.9_{\pm1.4}$ \\
        \texttt{yor\_Latn} & $29.4_{\pm0.6}$  & $29.1_{\pm1.6}$ & $40.6_{\pm1.3}$ & $41.4_{\pm0.7}$ & $38.8_{\pm1.2}$ & $42.0_{\pm0.8}$ \\
        \texttt{zho\_Hans} & $86.4_{\pm2.3}$  & $30.9_{\pm1.6}$ & $78.2_{\pm0.6}$ & $79.6_{\pm0.9}$ & $70.0_{\pm1.1}$ & $69.6_{\pm0.9}$ \\
        \texttt{zho\_Hant} & $85.0_{\pm2.2}$  & $33.2_{\pm1.3}$ & $75.5_{\pm1.1}$ & $76.8_{\pm0.5}$ & $51.5_{\pm0.3}$ & $57.9_{\pm0.5}$ \\
        \texttt{zsm\_Latn} & $68.6_{\pm9.2}$  & $29.0_{\pm0.5}$ & $79.7_{\pm1.4}$ & $81.5_{\pm0.8}$ & $73.4_{\pm0.4}$ & $73.9_{\pm1.1}$ \\
        \texttt{zul\_Latn} & $31.1_{\pm0.4}$  & $30.0_{\pm1.5}$ & $54.4_{\pm0.6}$ & $58.1_{\pm1.6}$ & $56.6_{\pm1.0}$ & $62.4_{\pm0.2}$ \\
         
        \bottomrule
    \end{tabular}
\end{adjustbox}
\caption{\textbf{Belebele (2/2).} We benchmark models on test portions of Belebele (cf. \S\ref{sec:experimental-setup}). S1 and S2 refer to self-supervised and task-specific stages of aligning NLLB with \llmvec (cf. \S\ref{sec:methodology}). FT denotes supervised fine-tuning. Reported performance is averaged over three seeds on model checkpoints that maximize performance on source-language (\sdev) validation splits. Subscripts denote std. deviation. Metric: accuracy.}
\label{tab:belebele-full-2}
\end{table*}

\begin{table*}[t!]
\rparagraph{Results by No. of Adaptation Steps}
\begin{center}
    \begin{tabular}{@{\extracolsep{\fill}} l!{\vrule width \arrayrulewidth}c!{\vrule width \arrayrulewidth}c!{\vrule width \arrayrulewidth}c!{\vrule width \arrayrulewidth}c}
        \toprule
         & \textbf{0} & \textbf{3,000} & \textbf{6,000} & \textbf{10,000} \\
        \hline
\textsc{xnli-en}          & $82.6_{\pm0.6}$ & $90.6_{\pm0.3}$ & $91.2_{\pm0.3}$ & $91.4_{\pm0.2}$ \\ \hline \hline
\textsc{amnli-aym}        & $60.0_{\pm0.7}$ & $62.8_{\pm2.4}$ & $62k0_{\pm1.2}$ & $61.3_{\pm0.8}$ \\
\textsc{amnli-gn}         & $65.7_{\pm0.4}$ & $68.5_{\pm0.5}$ & $69.5_{\pm0.8}$ & $69.7_{\pm1.3}$ \\
\textsc{amnli-quy}        & $60.7_{\pm1.5}$ & $61.8_{\pm0.8}$ & $61.4_{\pm2.4}$ & $61.5_{\pm1.9}$ \\
\textsc{kardeş-nlu-az}     & $79.3_{\pm0.9}$ & $84.1_{\pm0.1}$ & $83.7_{\pm1.1}$ & $83.1_{\pm0.7}$ \\
\textsc{kardeş-nlu-kk}     & $77.8_{\pm0.9}$ & $82.1_{\pm0.4}$ & $82.2_{\pm0.9}$ & $81.8_{\pm0.2}$ \\
\textsc{kardeş-nlu-ky}     & $77.9_{\pm0.6}$ & $81.9_{\pm0.1}$ & $81.6_{\pm0.3}$ & $81.4_{\pm0.6}$ \\
\textsc{kardeş-nlu-uz}     & $78.9_{\pm0.4}$ & $83.3_{\pm0.3}$ & $82.9_{\pm0.8}$ & $82.3_{\pm0.4}$ \\
\textsc{xnli-ar}          & $75.6_{\pm0.5}$ & $81.1_{\pm0.5}$ & $82.1_{\pm0.8}$ & $82.0_{\pm0.2}$ \\
\textsc{xnli-bg}          & $79.5_{\pm0.5}$ & $86.0_{\pm0.2}$ & $86.2_{\pm0.6}$ & $86.6_{\pm0.4}$ \\
\textsc{xnli-de}          & $79.1_{\pm0.2}$ & $85.6_{\pm0.4}$ & $85.8_{\pm0.4}$ & $85.8_{\pm0.2}$ \\
\textsc{xnli-el}          & $78.1_{\pm0.4}$ & $79.1_{\pm6.1}$ & $75.2_{\pm3.8}$ & $79.6_{\pm6.2}$ \\
\textsc{xnli-es}          & $79.8_{\pm0.4}$ & $86.9_{\pm0.3}$ & $87.3_{\pm0.3}$ & $87.5_{\pm0.3}$ \\
\textsc{xnli-fr}          & $78.8_{\pm0.5}$ & $86.2_{\pm0.4}$ & $86.6_{\pm0.3}$ & $86.8_{\pm0.3}$ \\
\textsc{xnli-hi}          & $72.7_{\pm0.3}$ & $76.4_{\pm0.5}$ & $76.6_{\pm1.0}$ & $76.9_{\pm0.6}$ \\
\textsc{xnli-ru}          & $76.9_{\pm0.2}$ & $83.4_{\pm0.7}$ & $83.7_{\pm0.6}$ & $83.9_{\pm0.1}$ \\
\textsc{xnli-sw}          & $73.5_{\pm0.2}$ & $79.5_{\pm0.4}$ & $79.6_{\pm0.4}$ & $79.6_{\pm0.4}$ \\
\textsc{xnli-th}          & $71.6_{\pm0.5}$ & $77.0_{\pm1.0}$ & $77.5_{\pm1.3}$ & $77.5_{\pm0.5}$ \\
\textsc{xnli-tr}          & $76.4_{\pm0.6}$ & $79.6_{\pm0.3}$ & $79.6_{\pm0.4}$ & $80.0_{\pm0.3}$ \\
\textsc{xnli-ur}          & $68.5_{\pm0.2}$ & $71.8_{\pm0.5}$ & $71.9_{\pm0.7}$ & $72.1_{\pm0.4}$ \\
\textsc{xnli-vi}          & $77.7_{\pm0.3}$ & $83.0_{\pm0.4}$ & $83.5_{\pm0.3}$ & $83.3_{\pm0.3}$ \\
\textsc{xnli-zh}          & $73.6_{\pm0.2}$ & $79.8_{\pm0.7}$ & $80.2_{\pm0.7}$ & $80.2_{\pm0.7}$ \\\hline
\textsc{avg}              & $74.4_{\pm0.3}$ & $79.1_{\pm0.2}$ & $79.0_{\pm0.4}$ & $79.2_{\pm0.2}$ \\
        \bottomrule
    \end{tabular}
\caption{\textbf{NLI by No. of Adaptation Steps.} We benchmark NLLB-\llmvec S1+S2 on test portions of NLI benchmarks (cf. \S\ref{sec:experimental-setup}) by number of adaptation steps in S1. S1 and S2 refer to self-supervised and task-specific stages of aligning NLLB with \llmvec (cf. \S\ref{sec:methodology}). FT denotes supervised fine-tuning. Reported performance is averaged over three seeds on model checkpoints that maximize performance on source-language (\sdev) validation splits. Subscripts denote std. deviation. Metric: accuracy.}
\label{tab:nli-by-step}
\end{center}
\end{table*}

\begin{table*}[t!]
\begin{center}
    \begin{tabular}{@{\extracolsep{\fill}} l!{\vrule width \arrayrulewidth}c!{\vrule width \arrayrulewidth}c!{\vrule width \arrayrulewidth}c!{\vrule width \arrayrulewidth}c}
        \toprule
         & \textbf{0} & \textbf{3,000} & \textbf{6,000} & \textbf{10,000} \\
        \hline
\textsc{eng} & $86.4_{\pm0.8}$ & $91.5_{\pm1.1}$ & $91.7_{\pm0.5}$ & $92.2_{\pm0.5}$ \\ \hline \hline
\textsc{ace} & $78.3_{\pm2.3}$ & $82.2_{\pm1.2}$ & $81.0_{\pm2.8}$ & $80.6_{\pm3.8}$ \\
\textsc{ban} & $75.2_{\pm1.9}$ & $75.0_{\pm2.2}$ & $72.4_{\pm3.1}$ & $72.9_{\pm3.7}$ \\
\textsc{bjn} & $82.3_{\pm1.7}$ & $82.7_{\pm0.7}$ & $82.7_{\pm1.3}$ & $82.6_{\pm2.0}$ \\
\textsc{bug} & $72.7_{\pm2.3}$ & $66.5_{\pm5.3}$ & $62.7_{\pm5.9}$ & $61.1_{\pm6.9}$ \\
\textsc{ind} & $85.5_{\pm1.6}$ & $88.6_{\pm0.8}$ & $89.3_{\pm1.1}$ & $89.3_{\pm0.7}$ \\
\textsc{jav} & $81.1_{\pm0.7}$ & $85.6_{\pm2.5}$ & $84.1_{\pm2.1}$ & $85.3_{\pm2.4}$ \\
\textsc{min} & $76.2_{\pm3.1}$ & $80.5_{\pm3.3}$ & $78.3_{\pm3.0}$ & $80.4_{\pm3.3}$ \\
\textsc{sun} & $80.8_{\pm1.9}$ & $83.9_{\pm2.0}$ & $82.6_{\pm3.1}$ & $83.2_{\pm3.1}$ \\ \hline
\textsc{avg} & $79.0_{\pm4.2}$ & $80.8_{\pm6.9}$ & $79.2_{\pm8.2}$ & $79.4_{\pm8.8}$ \\

        \bottomrule
    \end{tabular}
\caption{\textbf{NusaX by No. of Adaptation Steps.} We benchmark NLLB-\llmvec S1+S2 on test portions of NusaX (cf. \S\ref{sec:experimental-setup}) by number of adaptation steps in S1. S1 and S2 refer to self-supervised and task-specific stages of aligning NLLB with \llmvec (cf. \S\ref{sec:methodology}). FT denotes supervised fine-tuning. Reported performance is averaged over three seeds on model checkpoints that maximize performance on source-language (\sdev) validation splits. Subscripts denote std. deviation. Metric: accuracy.}
\label{tab:nusax-by-step}
\end{center}
\end{table*}

\begin{table*}[t!]
\begin{center}
{ \footnotesize
    \begin{tabular}{@{\extracolsep{\fill}} l!{\vrule width \arrayrulewidth}cccc}
        \toprule
         & \textbf{0} & \textbf{3,000} & \textbf{6,000} & \textbf{10,000} \\
        \hline
        \texttt{eng\_Latn} & $74.9_{\pm5.5}$  & $90.0_{\pm0.8}$ & $91.7_{\pm0.4}$ & $99.4_{\pm0.7}$ \\ \hline\hline
        AVG       & $51.3_{\pm2.8}$  & $61.8_{\pm0.2}$ & $62.2_{\pm0.2}$ & $62.6_{\pm0.5}$ \\ \hline
        
        \texttt{acm\_Arab} & $45.4_{\pm1.6}$ & $54.0_{\pm0.2}$ & $55.8_{\pm1.3}$ & $56.7_{\pm0.6}$\\
        \texttt{afr\_Latn} & $62.2_{\pm4.7}$ & $77.3_{\pm1.1}$ & $79.7_{\pm1.2}$ & $80.1_{\pm1.1}$\\
        \texttt{als\_Latn} & $54.3_{\pm5.3}$ & $71.0_{\pm0.9}$ & $71.7_{\pm0.4}$ & $72.9_{\pm1.0}$\\
        \texttt{amh\_Ethi} & $43.8_{\pm3.3}$ & $51.1_{\pm2.1}$ & $49.9_{\pm0.5}$ & $50.4_{\pm1.5}$\\
        \texttt{apc\_Arab} & $47.5_{\pm3.5}$ & $59.9_{\pm0.9}$ & $60.3_{\pm0.4}$ & $60.7_{\pm0.6}$\\
        \texttt{arb\_Arab} & $53.3_{\pm3.4}$ & $65.3_{\pm1.2}$ & $67.9_{\pm1.1}$ & $68.1_{\pm1.5}$\\
        \texttt{ars\_Arab} & $48.0_{\pm1.9}$ & $59.1_{\pm0.2}$ & $59.9_{\pm1.5}$ & $59.6_{\pm1.1}$\\
        \texttt{ary\_Arab} & $40.2_{\pm1.6}$ & $49.1_{\pm1.1}$ & $48.1_{\pm0.6}$ & $48.3_{\pm0.8}$\\
        \texttt{arz\_Arab} & $46.4_{\pm1.9}$ & $58.1_{\pm1.3}$ & $60.2_{\pm1.6}$ & $59.7_{\pm0.8}$\\
        \texttt{asm\_Beng} & $45.6_{\pm1.2}$ & $52.6_{\pm0.5}$ & $53.0_{\pm0.2}$ & $54.0_{\pm1.1}$\\
        \texttt{azj\_Latn} & $46.6_{\pm1.5}$ & $55.1_{\pm0.1}$ & $55.0_{\pm0.4}$ & $56.7_{\pm0.8}$\\
        \texttt{bam\_Latn} & $38.6_{\pm0.4}$ & $40.3_{\pm0.6}$ & $39.3_{\pm0.8}$ & $39.5_{\pm1.2}$\\
        \texttt{ben\_Beng} & $52.0_{\pm2.3}$ & $62.5_{\pm1.3}$ & $63.0_{\pm1.0}$ & $62.5_{\pm0.8}$\\
        \texttt{ben\_Latn} & $30.6_{\pm0.8}$ & $31.7_{\pm1.1}$ & $27.6_{\pm1.7}$ & $28.2_{\pm1.2}$\\
        \texttt{bod\_Tibt} & $30.5_{\pm0.1}$ & $34.5_{\pm0.3}$ & $33.5_{\pm0.7}$ & $33.9_{\pm1.1}$\\
        \texttt{bul\_Cyrl} & $61.0_{\pm3.8}$ & $76.5_{\pm0.3}$ & $76.3_{\pm0.4}$ & $77.9_{\pm0.1}$\\
        \texttt{cat\_Latn} & $63.7_{\pm3.4}$ & $78.9_{\pm0.6}$ & $80.9_{\pm0.5}$ & $82.3_{\pm0.9}$\\
        \texttt{ceb\_Latn} & $55.4_{\pm2.8}$ & $67.9_{\pm0.9}$ & $69.3_{\pm1.4}$ & $70.7_{\pm0.7}$\\
        \texttt{ces\_Latn} & $61.1_{\pm3.7}$ & $74.1_{\pm0.6}$ & $75.5_{\pm0.4}$ & $75.9_{\pm1.5}$\\
        \texttt{ckb\_Arab} & $48.9_{\pm0.8}$ & $58.4_{\pm1.0}$ & $58.7_{\pm1.0}$ & $59.8_{\pm1.2}$\\
        \texttt{dan\_Latn} & $66.2_{\pm4.0}$ & $80.8_{\pm1.4}$ & $82.8_{\pm0.4}$ & $83.5_{\pm0.7}$\\
        \texttt{deu\_Latn} & $64.4_{\pm4.7}$ & $78.4_{\pm0.7}$ & $79.1_{\pm0.0}$ & $78.1_{\pm0.9}$\\
        \texttt{ell\_Grek} & $51.7_{\pm3.2}$ & $64.6_{\pm1.2}$ & $67.3_{\pm0.8}$ & $67.5_{\pm0.2}$\\
        \texttt{est\_Latn} & $54.4_{\pm4.5}$ & $67.5_{\pm0.4}$ & $68.9_{\pm0.2}$ & $70.1_{\pm1.0}$\\
        \texttt{eus\_Latn} & $57.0_{\pm2.0}$ & $66.7_{\pm0.9}$ & $67.3_{\pm0.6}$ & $66.7_{\pm1.1}$\\
        \texttt{fin\_Latn} & $58.5_{\pm4.3}$ & $72.1_{\pm0.5}$ & $72.0_{\pm0.6}$ & $73.0_{\pm0.7}$\\
        \texttt{fra\_Latn} & $66.0_{\pm4.3}$ & $81.7_{\pm0.7}$ & $82.4_{\pm0.7}$ & $82.4_{\pm1.0}$\\
        \texttt{fuv\_Latn} & $29.3_{\pm0.4}$ & $29.7_{\pm0.2}$ & $28.6_{\pm0.9}$ & $28.1_{\pm0.6}$\\
        \texttt{gaz\_Latn} & $39.0_{\pm1.1}$ & $43.6_{\pm1.5}$ & $41.2_{\pm1.5}$ & $42.8_{\pm0.7}$\\
        \texttt{grn\_Latn} & $46.3_{\pm1.1}$ & $54.7_{\pm0.3}$ & $54.2_{\pm1.6}$ & $52.5_{\pm0.9}$\\
        \texttt{guj\_Gujr} & $47.0_{\pm2.1}$ & $55.3_{\pm0.6}$ & $55.9_{\pm0.8}$ & $55.9_{\pm1.7}$\\
        \texttt{hat\_Latn} & $53.5_{\pm2.5}$ & $67.1_{\pm1.0}$ & $67.8_{\pm0.9}$ & $67.4_{\pm0.9}$\\
        \texttt{hau\_Latn} & $48.6_{\pm2.9}$ & $60.8_{\pm1.5}$ & $61.4_{\pm0.4}$ & $62.1_{\pm0.4}$\\
        \texttt{heb\_Hebr} & $54.3_{\pm2.9}$ & $65.6_{\pm1.3}$ & $65.4_{\pm0.6}$ & $66.6_{\pm0.3}$\\
        \texttt{hin\_Deva} & $52.7_{\pm2.2}$ & $63.7_{\pm0.8}$ & $63.9_{\pm1.6}$ & $65.7_{\pm1.0}$\\
        \texttt{hrv\_Latn} & $59.4_{\pm4.9}$ & $75.0_{\pm1.1}$ & $76.6_{\pm0.3}$ & $77.4_{\pm0.5}$\\
        \texttt{hun\_Latn} & $57.2_{\pm4.4}$ & $71.5_{\pm0.3}$ & $71.6_{\pm0.4}$ & $71.6_{\pm0.5}$\\
        \texttt{hye\_Armn} & $48.6_{\pm2.8}$ & $58.8_{\pm0.7}$ & $57.8_{\pm1.0}$ & $58.6_{\pm0.6}$\\
        \texttt{ibo\_Latn} & $40.7_{\pm2.7}$ & $50.6_{\pm0.8}$ & $49.0_{\pm0.3}$ & $49.2_{\pm0.3}$\\
        \texttt{ilo\_Latn} & $53.4_{\pm2.2}$ & $64.5_{\pm1.1}$ & $65.1_{\pm0.8}$ & $66.7_{\pm1.3}$\\
        \texttt{ind\_Latn} & $66.4_{\pm4.4}$ & $81.2_{\pm0.3}$ & $81.7_{\pm1.1}$ & $82.5_{\pm0.6}$\\
        \texttt{isl\_Latn} & $50.7_{\pm3.1}$ & $63.4_{\pm0.6}$ & $65.1_{\pm0.8}$ & $65.0_{\pm1.2}$\\
        \texttt{ita\_Latn} & $64.9_{\pm4.7}$ & $80.3_{\pm1.1}$ & $81.4_{\pm0.9}$ & $82.4_{\pm0.5}$\\
        \texttt{jav\_Latn} & $60.7_{\pm3.7}$ & $73.0_{\pm0.1}$ & $73.4_{\pm0.4}$ & $74.2_{\pm0.5}$\\
        \texttt{jpn\_Jpan} & $53.5_{\pm3.8}$ & $68.0_{\pm0.2}$ & $67.6_{\pm1.2}$ & $67.1_{\pm0.9}$\\
        \texttt{kac\_Latn} & $37.4_{\pm0.6}$ & $39.4_{\pm1.2}$ & $40.2_{\pm1.1}$ & $40.9_{\pm1.2}$\\
        \texttt{kan\_Knda} & $49.9_{\pm3.6}$ & $56.4_{\pm0.9}$ & $56.5_{\pm1.1}$ & $56.9_{\pm1.3}$\\
        \texttt{kat\_Geor} & $44.4_{\pm2.4}$ & $51.2_{\pm0.7}$ & $50.3_{\pm0.9}$ & $51.1_{\pm0.5}$\\
        \texttt{kaz\_Cyrl} & $50.4_{\pm3.2}$ & $59.4_{\pm0.4}$ & $59.6_{\pm0.9}$ & $59.3_{\pm0.7}$\\
        \texttt{kea\_Latn} & $52.1_{\pm3.8}$ & $63.3_{\pm0.6}$ & $64.5_{\pm0.6}$ & $65.6_{\pm1.5}$\\
        \texttt{khk\_Cyrl} & $39.8_{\pm2.2}$ & $45.5_{\pm0.8}$ & $44.8_{\pm1.3}$ & $44.8_{\pm1.2}$\\
        \texttt{khm\_Khmr} & $40.1_{\pm0.5}$ & $51.3_{\pm0.8}$ & $48.8_{\pm1.5}$ & $51.6_{\pm0.8}$\\
        \texttt{kin\_Latn} & $48.0_{\pm2.6}$ & $55.9_{\pm0.9}$ & $57.0_{\pm0.3}$ & $57.1_{\pm0.8}$\\
        \texttt{kir\_Cyrl} & $51.1_{\pm2.8}$ & $57.4_{\pm1.1}$ & $58.9_{\pm1.2}$ & $58.7_{\pm0.8}$\\
        \texttt{kor\_Hang} & $52.3_{\pm3.8}$ & $63.7_{\pm0.6}$ & $63.0_{\pm0.9}$ & $62.3_{\pm0.7}$\\
        \texttt{lao\_Laoo} & $46.6_{\pm2.9}$ & $58.5_{\pm0.5}$ & $58.6_{\pm0.5}$ & $58.4_{\pm1.5}$\\
        \texttt{lin\_Latn} & $47.4_{\pm3.6}$ & $56.4_{\pm0.6}$ & $56.6_{\pm0.4}$ & $57.0_{\pm0.8}$\\
        \texttt{lit\_Latn} & $58.0_{\pm4.8}$ & $69.7_{\pm1.1}$ & $70.1_{\pm0.4}$ & $72.3_{\pm0.5}$\\
        \texttt{lug\_Latn} & $40.4_{\pm2.1}$ & $46.1_{\pm0.1}$ & $46.0_{\pm1.2}$ & $47.2_{\pm0.3}$\\
        \texttt{luo\_Latn} & $39.4_{\pm2.5}$ & $46.5_{\pm0.6}$ & $46.9_{\pm0.1}$ & $45.4_{\pm0.9}$\\
        \texttt{lvs\_Latn} & $57.3_{\pm3.7}$ & $69.3_{\pm0.5}$ & $69.9_{\pm0.2}$ & $70.6_{\pm1.3}$\\
         
        \bottomrule
    \end{tabular}
}
\end{center}
\caption{\textbf{Belebele by No. of Adaptation Steps (1/2).} We benchmark NLLB-\llmvec S1+S2 on test portions of Belebele (cf. \S\ref{sec:experimental-setup}) by number of adaptation steps in S1. S1 and S2 refer to self-supervised and task-specific stages of aligning NLLB with \llmvec (cf. \S\ref{sec:methodology}). FT denotes supervised fine-tuning. Reported performance is averaged over three seeds on model checkpoints that maximize performance on source-language (\sdev) validation splits. Subscripts denote std. deviation. Metric: accuracy.}
\label{tab:belebele-by-step-1}
\end{table*}

\begin{table*}[t!]
\begin{center}
{\footnotesize
    \begin{tabular}{@{\extracolsep{\fill}} l!{\vrule width \arrayrulewidth}cccc}
        \toprule
         & \textbf{0} & \textbf{3,000} & \textbf{6,000} & \textbf{10,000} \\
        \hline
        \texttt{eng\_Latn} & $74.9_{\pm5.5}$  & $90.0_{\pm0.8}$ & $91.7_{\pm0.4}$ & $99.4_{\pm0.7}$ \\ \hline\hline
        AVG       & $51.3_{\pm2.8}$  & $61.8_{\pm0.2}$ & $62.2_{\pm0.2}$ & $62.6_{\pm0.5}$ \\ \hline
        
        \texttt{mal\_Mlym} & $44.0_{\pm2.9}$ & $52.1_{\pm1.1}$ & $50.6_{\pm0.9}$ & $49.9_{\pm0.4}$ \\
        \texttt{mar\_Deva} & $51.7_{\pm3.7}$ & $60.6_{\pm0.9}$ & $60.3_{\pm1.4}$ & $60.8_{\pm0.5}$ \\
        \texttt{mkd\_Cyrl} & $59.4_{\pm3.1}$ & $72.2_{\pm1.2}$ & $73.4_{\pm0.5}$ & $74.3_{\pm0.6}$ \\
        \texttt{mlt\_Latn} & $52.1_{\pm1.5}$ & $65.7_{\pm0.7}$ & $66.5_{\pm0.5}$ & $67.7_{\pm1.7}$ \\
        \texttt{mri\_Latn} & $41.2_{\pm2.5}$ & $45.8_{\pm0.3}$ & $46.7_{\pm0.6}$ & $47.2_{\pm1.4}$ \\
        \texttt{mya\_Mymr} & $40.6_{\pm1.8}$ & $47.3_{\pm0.6}$ & $46.6_{\pm0.7}$ & $47.4_{\pm0.5}$ \\
        \texttt{nld\_Latn} & $64.1_{\pm3.9}$ & $79.1_{\pm0.4}$ & $80.7_{\pm0.4}$ & $81.0_{\pm0.6}$ \\
        \texttt{nob\_Latn} & $65.2_{\pm4.1}$ & $81.8_{\pm1.2}$ & $83.9_{\pm0.7}$ & $84.6_{\pm0.2}$ \\
        \texttt{npi\_Deva} & $51.5_{\pm2.8}$ & $61.1_{\pm1.1}$ & $61.2_{\pm0.4}$ & $60.8_{\pm1.6}$ \\
        \texttt{nso\_Latn} & $49.6_{\pm4.3}$ & $58.0_{\pm0.8}$ & $60.4_{\pm0.9}$ & $60.0_{\pm1.5}$ \\
        \texttt{nya\_Latn} & $44.4_{\pm1.5}$ & $52.9_{\pm1.2}$ & $54.5_{\pm1.4}$ & $54.5_{\pm1.4}$ \\
        \texttt{ory\_Orya} & $48.2_{\pm1.4}$ & $57.1_{\pm0.8}$ & $57.0_{\pm1.1}$ & $57.1_{\pm1.2}$ \\
        \texttt{pan\_Guru} & $46.6_{\pm2.0}$ & $55.2_{\pm1.4}$ & $55.0_{\pm1.1}$ & $56.5_{\pm2.1}$ \\
        \texttt{pbt\_Arab} & $44.2_{\pm2.3}$ & $50.6_{\pm1.3}$ & $48.7_{\pm0.8}$ & $49.9_{\pm1.4}$ \\
        \texttt{pes\_Arab} & $60.1_{\pm5.5}$ & $72.4_{\pm0.5}$ & $71.6_{\pm0.4}$ & $71.4_{\pm0.3}$ \\
        \texttt{plt\_Latn} & $52.3_{\pm2.8}$ & $64.9_{\pm0.7}$ & $64.1_{\pm0.5}$ & $64.9_{\pm0.7}$ \\
        \texttt{pol\_Latn} & $55.7_{\pm3.9}$ & $69.0_{\pm0.4}$ & $70.7_{\pm0.8}$ & $71.0_{\pm0.8}$ \\
        \texttt{por\_Latn} & $68.2_{\pm4.8}$ & $83.2_{\pm0.5}$ & $83.8_{\pm0.7}$ & $84.1_{\pm0.7}$ \\
        \texttt{ron\_Latn} & $62.1_{\pm4.3}$ & $77.3_{\pm0.9}$ & $78.0_{\pm1.0}$ & $79.1_{\pm1.0}$ \\
        \texttt{rus\_Cyrl} & $61.6_{\pm3.6}$ & $76.2_{\pm0.6}$ & $78.1_{\pm1.0}$ & $79.1_{\pm1.2}$ \\
        \texttt{shn\_Mymr} & $33.3_{\pm1.4}$ & $36.4_{\pm1.6}$ & $36.4_{\pm0.8}$ & $37.3_{\pm0.5}$ \\
        \texttt{sin\_Sinh} & $38.4_{\pm2.3}$ & $43.9_{\pm1.3}$ & $44.1_{\pm1.7}$ & $43.8_{\pm1.6}$ \\
        \texttt{slk\_Latn} & $60.5_{\pm4.6}$ & $73.3_{\pm0.2}$ & $74.9_{\pm0.3}$ & $76.0_{\pm0.8}$ \\
        \texttt{slv\_Latn} & $59.9_{\pm4.2}$ & $74.9_{\pm0.7}$ & $76.2_{\pm0.3}$ & $76.0_{\pm0.6}$ \\
        \texttt{sna\_Latn} & $48.8_{\pm2.2}$ & $61.1_{\pm1.0}$ & $61.4_{\pm0.4}$ & $61.1_{\pm0.9}$ \\
        \texttt{snd\_Arab} & $48.6_{\pm2.4}$ & $57.6_{\pm1.4}$ & $57.1_{\pm1.2}$ & $56.6_{\pm0.3}$ \\
        \texttt{som\_Latn} & $43.3_{\pm2.1}$ & $52.0_{\pm1.5}$ & $51.4_{\pm0.9}$ & $51.2_{\pm1.1}$ \\
        \texttt{sot\_Latn} & $45.6_{\pm3.7}$ & $56.6_{\pm0.1}$ & $57.4_{\pm1.4}$ & $57.9_{\pm1.4}$ \\
        \texttt{spa\_Latn} & $67.0_{\pm3.6}$ & $84.6_{\pm0.8}$ & $85.3_{\pm0.2}$ & $85.6_{\pm0.4}$ \\
        \texttt{srp\_Cyrl} & $60.0_{\pm3.9}$ & $72.8_{\pm1.0}$ & $74.4_{\pm0.8}$ & $75.5_{\pm1.0}$ \\
        \texttt{ssw\_Latn} & $44.2_{\pm3.9}$ & $51.4_{\pm0.2}$ & $51.7_{\pm0.2}$ & $52.4_{\pm2.1}$ \\
        \texttt{sun\_Latn} & $54.8_{\pm3.2}$ & $68.8_{\pm0.9}$ & $68.9_{\pm0.4}$ & $70.1_{\pm0.6}$ \\
        \texttt{swe\_Latn} & $66.6_{\pm4.6}$ & $81.3_{\pm0.7}$ & $81.9_{\pm0.1}$ & $82.8_{\pm0.4}$ \\
        \texttt{swh\_Latn} & $60.3_{\pm4.2}$ & $74.9_{\pm0.4}$ & $74.6_{\pm0.5}$ & $75.5_{\pm0.9}$ \\
        \texttt{tam\_Taml} & $47.2_{\pm3.9}$ & $55.7_{\pm1.7}$ & $55.6_{\pm0.9}$ & $54.8_{\pm0.9}$ \\
        \texttt{tel\_Telu} & $47.0_{\pm2.6}$ & $56.0_{\pm0.6}$ & $54.3_{\pm1.2}$ & $54.6_{\pm1.0}$ \\
        \texttt{tgk\_Cyrl} & $45.6_{\pm3.1}$ & $55.6_{\pm0.2}$ & $55.4_{\pm0.5}$ & $54.7_{\pm0.6}$ \\
        \texttt{tgl\_Latn} & $58.0_{\pm3.1}$ & $71.8_{\pm0.3}$ & $72.1_{\pm0.7}$ & $72.2_{\pm1.3}$ \\
        \texttt{tha\_Thai} & $46.5_{\pm1.9}$ & $58.0_{\pm0.6}$ & $58.1_{\pm0.8}$ & $58.8_{\pm0.6}$ \\
        \texttt{tir\_Ethi} & $40.1_{\pm2.0}$ & $44.0_{\pm0.7}$ & $44.9_{\pm1.5}$ & $44.6_{\pm1.8}$ \\
        \texttt{tsn\_Latn} & $48.3_{\pm3.8}$ & $55.9_{\pm1.0}$ & $55.4_{\pm0.9}$ & $56.6_{\pm0.6}$ \\
        \texttt{tso\_Latn} & $54.9_{\pm1.8}$ & $64.3_{\pm0.6}$ & $63.7_{\pm0.2}$ & $64.5_{\pm1.7}$ \\
        \texttt{tur\_Latn} & $55.3_{\pm3.2}$ & $62.4_{\pm0.7}$ & $63.7_{\pm0.8}$ & $64.0_{\pm1.1}$ \\
        \texttt{ukr\_Cyrl} & $57.0_{\pm3.1}$ & $72.3_{\pm0.2}$ & $72.9_{\pm1.1}$ & $74.1_{\pm0.8}$ \\
        \texttt{urd\_Arab} & $53.8_{\pm2.1}$ & $63.2_{\pm1.2}$ & $63.5_{\pm0.7}$ & $63.6_{\pm1.6}$ \\
        \texttt{uzn\_Latn} & $49.8_{\pm2.3}$ & $60.5_{\pm0.9}$ & $61.3_{\pm0.3}$ & $61.0_{\pm0.8}$ \\
        \texttt{vie\_Latn} & $60.3_{\pm3.4}$ & $75.5_{\pm0.2}$ & $75.2_{\pm1.0}$ & $76.4_{\pm0.3}$ \\
        \texttt{war\_Latn} & $55.9_{\pm4.0}$ & $69.6_{\pm0.4}$ & $69.0_{\pm1.0}$ & $68.6_{\pm1.1}$ \\
        \texttt{wol\_Latn} & $34.7_{\pm1.5}$ & $38.6_{\pm1.5}$ & $37.0_{\pm0.8}$ & $36.9_{\pm0.7}$ \\
        \texttt{xho\_Latn} & $49.7_{\pm2.5}$ & $59.4_{\pm1.1}$ & $60.6_{\pm0.1}$ & $61.2_{\pm0.8}$ \\
        \texttt{yor\_Latn} & $36.1_{\pm1.7}$ & $41.6_{\pm0.7}$ & $41.1_{\pm0.9}$ & $41.4_{\pm0.7}$ \\
        \texttt{zho\_Hans} & $65.3_{\pm3.7}$ & $78.4_{\pm0.7}$ & $79.9_{\pm0.5}$ & $79.6_{\pm0.9}$ \\
        \texttt{zho\_Hant} & $63.9_{\pm3.6}$ & $77.4_{\pm0.4}$ & $77.6_{\pm0.7}$ & $76.8_{\pm0.5}$ \\
        \texttt{zsm\_Latn} & $66.1_{\pm3.8}$ & $80.1_{\pm1.0}$ & $80.7_{\pm1.5}$ & $81.5_{\pm0.8}$ \\
        \texttt{zul\_Latn} & $47.1_{\pm2.3}$ & $56.2_{\pm0.9}$ & $56.7_{\pm0.9}$ & $58.1_{\pm1.6}$ \\
         
        \bottomrule
    \end{tabular}
}
\end{center}
\caption{\textbf{Belebele by No. of Adaptation Steps (2/2).} We benchmark NLLB-\llmvec S1+S2 on test portions of Belebele (cf. \S\ref{sec:experimental-setup}) by number of adaptation steps in S1. S1 and S2 refer to self-supervised and task-specific stages of aligning NLLB with \llmvec (cf. \S\ref{sec:methodology}). FT denotes supervised fine-tuning. Reported performance is averaged over three seeds on model checkpoints that maximize performance on source-language (\sdev) validation splits. Subscripts denote std. deviation. Metric: accuracy.}
\label{tab:belebele-by-step-2}
\end{table*}

\begin{table*}[t!]
\rparagraph{NLLB-GPT-2}\\

\begin{adjustbox}{width=\textwidth,center}
    \begin{tabular}{@{\extracolsep{\fill}} l!{\vrule width \arrayrulewidth}c    !{\vrule width \arrayrulewidth}!{\vrule width \arrayrulewidth}    c!{\vrule width \arrayrulewidth}c!{\vrule width \arrayrulewidth}c!{\vrule width \arrayrulewidth}c!{\vrule width \arrayrulewidth}c!{\vrule width \arrayrulewidth}c!{\vrule width \arrayrulewidth}c!{\vrule width \arrayrulewidth}c!{\vrule width \arrayrulewidth}c!{\vrule width \arrayrulewidth}c!{\vrule width \arrayrulewidth}c!{\vrule width \arrayrulewidth}c!{\vrule width \arrayrulewidth}c!{\vrule width \arrayrulewidth}c!{\vrule width \arrayrulewidth}c}
        \toprule
         & \textbf{\textsc{EN}} & \textbf{\textsc{AR}} & \textbf{\textsc{BG}} & \textbf{\textsc{DE}} & \textbf{\textsc{EL}} & \textbf{\textsc{ES}} & \textbf{\textsc{FR}} & \textbf{\textsc{HI}} & \textbf{\textsc{RU}} & \textbf{\textsc{SW}} & \textbf{\textsc{TH}} & \textbf{\textsc{TR}} & \textbf{\textsc{UR}} & \textbf{\textsc{VI}} & \textbf{\textsc{ZH}}  & \textbf{\textsc{AVG}} \\
        \hline
        \rowcolor{gray!10} \multicolumn{17}{l}{\rule{0pt}{2.5ex} \textit{Zero-Shot Cross-Lingual Transfer: Fine-tune multilingual model on English training set}}  \\ 
        \hline
        NLLB-GPT-2 FT & $82.2_{\pm0.1}$ & $75.0_{\pm0.4}$ & $78.3_{\pm0.6}$ & $77.1_{\pm1.0}$ & $75.6_{\pm0.5}$ & $78.8_{\pm0.4}$ & $77.9_{\pm0.8}$ & $71.2_{\pm0.5}$ & $75.6_{\pm0.4}$ & $73.1_{\pm0.6}$ & $71.1_{\pm0.8}$ & $74.0_{\pm0.8}$ & $68.4_{\pm0.9}$ & $76.7_{\pm0.4}$ & $73.4_{\pm0.6}$ & $74.7_{\pm0.5}$ \\
        \hline
        \rowcolor{gray!10} \multicolumn{17}{l}{\rule{0pt}{2.5ex} \textit{Translate-Test: Translate test data to English}} \\ 
        \hline
        GPT-2 NLLB-600M  & $00.0_{\pm0.0}$ & $73.9_{\pm0.1}$ & $76.3_{\pm0.3}$ & $77.4_{\pm0.1}$ & $77.5_{\pm0.6}$ & $78.7_{\pm0.2}$ & $78.4_{\pm0.5}$ & $71.3_{\pm0.4}$ & $73.8_{\pm0.2}$ & $69.1_{\pm0.4}$ & $69.0_{\pm0.3}$ & $74.1_{\pm0.5}$ & $65.6_{\pm0.1}$ & $74.9_{\pm0.3}$ & $71.0_{\pm0.5}$ & $73.6_{\pm0.1}$ \\
        GPT-2 NLLB-3B    & $00.0_{\pm0.3}$ & $74.9_{\pm0.5}$ & $78.1_{\pm0.1}$ & $78.9_{\pm0.2}$ & $77.9_{\pm0.3}$ & $79.8_{\pm0.2}$ & $79.4_{\pm0.1}$ & $72.6_{\pm0.3}$ & $76.0_{\pm0.4}$ & $69.9_{\pm0.3}$ & $71.0_{\pm0.1}$ & $76.6_{\pm0.6}$ & $66.0_{\pm0.4}$ & $75.9_{\pm0.4}$ & $73.7_{\pm1.0}$ & $75.1_{\pm0.2}$ \\
        \bottomrule
    \end{tabular}
\end{adjustbox}
\caption{\textbf{NLLB-GPT-2 on XNLI (1/2).} We benchmark models on test portions of XNLI (cf. \S\ref{sec:experimental-setup}). S1 refers to the self-supervised stage of aligning NLLB with GPT-2 (cf. \S\ref{sec:methodology}). FT denotes supervised fine-tuning. Reported performance is averaged over three seeds on model checkpoints that maximize performance on \textbf{source-language (\sdev) validation splits}. Subscripts denote std. deviation. Metric: accuracy.}
\label{tab:xnli-src-dev-gpt2}
\end{table*}

\begin{table*}[t!]
\begin{adjustbox}{width=\textwidth,center}
    \begin{tabular}{@{\extracolsep{\fill}} l!{\vrule width \arrayrulewidth}c    !{\vrule width \arrayrulewidth}!{\vrule width \arrayrulewidth}    c!{\vrule width \arrayrulewidth}c!{\vrule width \arrayrulewidth}c!{\vrule width \arrayrulewidth}c!{\vrule width \arrayrulewidth}c!{\vrule width \arrayrulewidth}c!{\vrule width \arrayrulewidth}c!{\vrule width \arrayrulewidth}c!{\vrule width \arrayrulewidth}c!{\vrule width \arrayrulewidth}c!{\vrule width \arrayrulewidth}c!{\vrule width \arrayrulewidth}c!{\vrule width \arrayrulewidth}c!{\vrule width \arrayrulewidth}c!{\vrule width \arrayrulewidth}c}
        \toprule
         & \textbf{\textsc{EN}} & \textbf{\textsc{AR}} & \textbf{\textsc{BG}} & \textbf{\textsc{DE}} & \textbf{\textsc{EL}} & \textbf{\textsc{ES}} & \textbf{\textsc{FR}} & \textbf{\textsc{HI}} & \textbf{\textsc{RU}} & \textbf{\textsc{SW}} & \textbf{\textsc{TH}} & \textbf{\textsc{TR}} & \textbf{\textsc{UR}} & \textbf{\textsc{VI}} & \textbf{\textsc{ZH}} & \textbf{\textsc{AVG}} \\
        \hline
        \rowcolor{gray!10} \multicolumn{17}{l}{\rule{0pt}{2.5ex} \textit{Zero-Shot Cross-Lingual Transfer: Fine-tune multilingual model on English training set}}  \\ 
        \hline
        NLLB-GPT-2 S1+FT & $82.2_{\pm0.1}$ & $75.0_{\pm0.4}$ & $78.4_{\pm0.5}$ & $77.2_{\pm0.9}$ & $75.8_{\pm0.6}$ & $78.8_{\pm0.4}$ & $78.0_{\pm0.8}$ & $71.3_{\pm0.5}$ & $75.7_{\pm0.4}$ & $73.1_{\pm0.6}$ & $71.1_{\pm0.8}$ & $74.2_{\pm0.7}$ & $68.4_{\pm0.9}$ & $76.7_{\pm0.3}$ & $73.5_{\pm0.4}$ & $74.8_{\pm0.5}$  \\
        \hline
        \rowcolor{gray!10} \multicolumn{17}{l}{\rule{0pt}{2.5ex} \textit{Translate-Test: Translate test data to English}} \\ 
        \hline
        GPT-2 NLLB-600M  & $85.0_{\pm0.1}$ & $74.1_{\pm0.2}$ & $76.8_{\pm0.6}$ & $77.7_{\pm0.1}$ & $77.8_{\pm0.3}$ & $79.1_{\pm0.3}$ & $78.6_{\pm0.5}$ & $71.7_{\pm0.2}$ & $74.0_{\pm0.1}$ & $69.6_{\pm0.2}$ & $69.4_{\pm0.2}$ & $74.4_{\pm0.4}$ & $65.9_{\pm0.4}$ & $75.1_{\pm0.3}$ & $71.7_{\pm0.6}$ & $74.0_{\pm0.1}$ \\
        GPT-2 NLLB-3B    & $85.0_{\pm0.1}$ & $75.1_{\pm0.4}$ & $78.6_{\pm0.3}$ & $79.1_{\pm0.3}$ & $78.3_{\pm0.3}$ & $80.1_{\pm0.3}$ & $79.6_{\pm0.3}$ & $72.9_{\pm0.1}$ & $76.4_{\pm0.1}$ & $70.5_{\pm0.3}$ & $71.4_{\pm0.3}$ & $76.7_{\pm0.4}$ & $66.4_{\pm0.1}$ & $76.1_{\pm0.2}$ & $74.4_{\pm0.4}$ & $75.4_{\pm0.0}$  \\
        \bottomrule
    \end{tabular}
\end{adjustbox}
\caption{\textbf{NLLB-GPT-2 on XNLI (2/2).} We benchmark models on test portions of XNLI (cf. \S\ref{sec:experimental-setup}). S1 refers to the self-supervised stage of aligning NLLB with GPT-2 (cf. \S\ref{sec:methodology}). FT denotes supervised fine-tuning. Reported performance is averaged over three seeds on model checkpoints that maximize performance on \textbf{per target-language (\tdev) validation splits}. Subscripts denote std. deviation. Metric: accuracy.}
\label{tab:xnli-trg-dev-gpt2}
\end{table*}

\begin{table*}[t!]
\begin{adjustbox}{width=\textwidth,center}
    \begin{tabular}{@{\extracolsep{\fill}} l!{\vrule width \arrayrulewidth}cc!{\vrule width \arrayrulewidth}cc!{\vrule width \arrayrulewidth}cc!{\vrule width \arrayrulewidth}cc!{\vrule width \arrayrulewidth}    !{\vrule width \arrayrulewidth}cc!{\vrule width \arrayrulewidth}cc!{\vrule width \arrayrulewidth}cc!{\vrule width \arrayrulewidth}cc!{\vrule width \arrayrulewidth}cc}
        \toprule
        \multicolumn{1}{c}{} & \multicolumn{2}{c}{\textbf{\textsc{AYM}}} & \multicolumn{2}{c}{\textbf{\textsc{GN}}} & \multicolumn{2}{c}{\textbf{\textsc{QUY}}} & \multicolumn{2}{c}{\textbf{\textsc{AVG}}} & \multicolumn{2}{c}{\textbf{\textsc{AZ}}} & \multicolumn{2}{c}{\textbf{\textsc{KK}}} & \multicolumn{2}{c}{\textbf{\textsc{KY}}} & \multicolumn{2}{c}{\textbf{\textsc{UZ}}} & \multicolumn{2}{c}{\textbf{\textsc{AVG}}} \\
        \cmidrule(lr){2-3} \cmidrule(lr){4-5} \cmidrule(lr){6-7} \cmidrule(lr){8-9} \cmidrule(lr){10-11} \cmidrule(lr){12-13} \cmidrule(lr){14-15} \cmidrule(lr){16-17} \cmidrule(lr){18-19}
        & \sdev & \tdev & \sdev & \tdev & \sdev & \tdev & \sdev & \tdev & \sdev & \tdev & \sdev & \tdev & \sdev & \tdev & \sdev & \tdev & \sdev & \tdev \\
        \hline
        \rowcolor{gray!10} \multicolumn{19}{l}{\rule{0pt}{2.5ex} \textit{Zero-Shot Cross-Lingual Transfer: Fine-tune multilingual model on English training set}}  \\ 
        \hline
        NLLB-GPT-2 S1+FT & $59.7_{\pm0.5}$ & $60.4_{\pm0.4}$ & $66.8_{\pm0.9}$ & $67.8_{\pm0.7}$ & $60.0_{\pm0.8}$ & $61.2_{\pm1.1}$ & $62.2_{\pm0.6}$ & $63.2_{\pm0.6}$ & $77.4_{\pm0.4}$ & $77.7_{\pm0.7}$ & $75.3_{\pm0.3}$ & $75.5_{\pm0.1}$ & $75.6_{\pm0.3}$ & $75.7_{\pm0.4}$ & $76.6_{\pm0.6}$ & $77.1_{\pm0.6}$ & $76.2{\pm0.2}$ & $76.5_{\pm0.3}$\\
        \hline
        \rowcolor{gray!10} \multicolumn{19}{l}{\rule{0pt}{2.5ex} \textit{Translate-Test: Translate test data to English}} \\ 
        \hline
        GPT-2 NLLB-600M & $50.3_{\pm1.1}$ & $52.1_{\pm1.1}$ & $58.0_{\pm0.9}$ & $59.0_{\pm0.6}$ & $54.0_{\pm0.5}$ & $54.4_{\pm0.1}$ & $54.1_{\pm0.5}$ & $55.2_{\pm0.4}$ & $77.2_{\pm0.2}$ & $77.4_{\pm0.5}$ & $73.2_{\pm0.1}$ & $73.7_{\pm0.4}$ & $73.5_{\pm0.4}$ & $74.3_{\pm0.0}$ & $75.0_{\pm0.5}$ & $75.7_{\pm0.7}$ & $74.7_{\pm0.0}$ & $75.3_{\pm0.2}$  \\
        GPT-2 NLLB-3B   & $46.2_{\pm1.0}$ & $47.9_{\pm0.6}$ & $59.7_{\pm1.5}$ & $61.6_{\pm0.3}$ & $51.9_{\pm1.0}$ & $54.0_{\pm0.3}$ & $52.6_{\pm0.5}$ & $54.5_{\pm0.2}$ & $78.9_{\pm0.3}$ & $79.3_{\pm0.4}$ & $75.0_{\pm0.4}$ & $75.5_{\pm0.4}$ & $71.7_{\pm0.4}$ & $72.6_{\pm0.4}$ & $75.9_{\pm0.4}$ & $76.3_{\pm0.3}$ & $75.4_{\pm0.4}$ & $75.9_{\pm0.3}$ \\
        \bottomrule
    \end{tabular}
\end{adjustbox}
\caption{ \textbf{NLLB-GPT-2 on AmNLI \& Kardeş-NLU.} We benchmark models on test portions of AmNLI and Kardeş-NLU (cf. \S\ref{sec:experimental-setup}). S1 refers to the self-supervised stage of aligning NLLB with GPT-2 (cf. \S\ref{sec:methodology}). FT denotes supervised fine-tuning. Reported performance is averaged over three seeds on model checkpoints that maximize performance on source-language (\sdev) and per target-language (\tdev) validation splits. Subscripts denote std. deviation. Metric: accuracy. }
\label{tab:amnli-kardesnlu-gpt2}
\end{table*}

\end{document}